\documentclass[journal]{IEEEtran}

\usepackage{tikz}
\usepackage{xifthen}

\pgfdeclarelayer{bg}    
\pgfsetlayers{bg,main}  
\definecolor{beige}{RGB}{245, 245, 220}
\definecolor{darkgrey}{RGB}{75, 75, 75}
\definecolor{lightgrey}{RGB}{250, 250, 250}
\newcommand*\circled[1]{\tikz[baseline=(char.base)]{
            \node[shape=circle,draw,inner sep=1pt] (char) {#1};}}
\newcommand*\smallcircled[1]{\tikz[baseline=(char.base)]{
            \node[shape=circle,draw,inner sep=0pt,minimum width=3.5mm] (char) {#1};}}

\usetikzlibrary{calc, arrows, fit, positioning, patterns, decorations.pathreplacing, shapes}
\tikzstyle{dash} = [dashed, -latex,>=latex]
\tikzstyle{line} = [draw, -latex,>=latex]
\tikzstyle{box} = [draw, minimum size=.8cm]
\tikzstyle{roundbox} = [draw, circle, inner sep=0pt, minimum size=3mm]
\tikzstyle{clamped} = [draw, fill=black, minimum size=0.15cm]
\tikzstyle{msgcircle} = [shape=circle, draw, inner sep=0pt, minimum size=5mm, fill=white]
\tikzstyle{darkmsgcircle} = [shape=circle, draw, inner sep=0pt, minimum size=5mm, fill=darkgrey, text=white, font=\bfseries]
\tikzstyle{msgdoublecircle} = [shape=circle, double, double distance=1.5pt, draw, inner sep=0pt, minimum size=5mm, fill=white]
\tikzstyle{darkmsgdoublecircle} = [shape=circle, double, double distance=1.5pt, draw, inner sep=0pt, minimum size=5mm, fill=darkgrey, text=white, font=\bfseries]

\newcommand{\brackets}[4]{
	\draw[dotted, rounded corners=0.3cm, line width = 1pt] ($({#1},{#3})+(-0.3,0)$) -- ($({#1},{#3})+(-0.3,0.3)$) -- ($({#2},{#3})+(0.3,0.3)$) -- ($({#2},{#3})+(0.3,0)$);
	\draw[dotted, rounded corners=0.3cm, line width = 1pt] ($({#1},{#4})+(-0.3,0)$) -- ($({#1},{#4})+(-0.3,-0.3)$) -- ($({#2},{#4})+(0.3,-0.3)$) -- ($({#2},{#4})+(0.3,0)$);}

\newcommand{\msg}[6]{
      \ifthenelse{\isin{#1}{left} \AND \isin{#2}{down}}{
            \coordinate (anchor) at ($({#3})!{#5}!({#4})$);
            \node[msgcircle, xshift=-5.5mm] at (anchor) {#6};
            \node[xshift=-1.5mm] at (anchor) {$\downarrow$};
      }{}
      \ifthenelse{\isin{#1}{right} \AND \isin{#2}{down}}{
            \coordinate (anchor) at ($({#3})!{#5}!({#4})$);
            \node[msgcircle, xshift=5.5mm] at (anchor) {#6};
            \node[xshift=1.5mm] at (anchor) {$\downarrow$};
      }{}

      \ifthenelse{\isin{#1}{down} \AND \isin{#2}{right}}{
            \coordinate (anchor) at ($({#3})!{#5}!({#4})$);
            \node[msgcircle, yshift=-6.0mm] at (anchor) {#6};
            \node[yshift=-2.0mm] at (anchor) {$\rightarrow$};
      }{}
      \ifthenelse{\isin{#1}{up} \AND \isin{#2}{right}}{
            \coordinate (anchor) at ($({#3})!{#5}!({#4})$);
            \node[msgcircle, yshift=6.0mm] at (anchor) {#6};
            \node[yshift=2.0mm] at (anchor) {$\rightarrow$};
      }{}

      \ifthenelse{\isin{#1}{down} \AND \isin{#2}{left}}{
            \coordinate (anchor) at ($({#3})!{#5}!({#4})$);
            \node[msgcircle, yshift=-6.0mm] at (anchor) {#6};
            \node[yshift=-2.0mm] at (anchor) {$\leftarrow$};
      }{}
      \ifthenelse{\isin{#1}{up} \AND \isin{#2}{left}}{
            \coordinate (anchor) at ($({#3})!{#5}!({#4})$);
            \node[msgcircle, yshift=6.0mm] at (anchor) {#6};
            \node[yshift=2.0mm] at (anchor) {$\leftarrow$};
      }{}

      \ifthenelse{\isin{#1}{left} \AND \isin{#2}{up}}{
            \coordinate (anchor) at ($({#3})!{#5}!({#4})$);
            \node[msgcircle, xshift=-5.5mm] at (anchor) {#6};
            \node[xshift=-1.5mm] at (anchor) {$\uparrow$};
      }{}
      \ifthenelse{\isin{#1}{right} \AND \isin{#2}{up}}{
            \coordinate (anchor) at ($({#3})!{#5}!({#4})$);
            \node[msgcircle, xshift=5.5mm] at (anchor) {#6};
            \node[xshift=1.5mm] at (anchor) {$\uparrow$};
      }{}
}

\newcommand{\darkmsg}[6]{
      \ifthenelse{\isin{#1}{left} \AND \isin{#2}{down}}{
            \coordinate (anchor) at ($({#3})!{#5}!({#4})$);
            \node[darkmsgcircle, xshift=-5.5mm] at (anchor) {#6};
            \node[xshift=-1.5mm] at (anchor) {$\downarrow$};
      }{}
      \ifthenelse{\isin{#1}{right} \AND \isin{#2}{down}}{
            \coordinate (anchor) at ($({#3})!{#5}!({#4})$);
            \node[darkmsgcircle, xshift=5.5mm] at (anchor) {#6};
            \node[xshift=1.5mm] at (anchor) {$\downarrow$};
      }{}

      \ifthenelse{\isin{#1}{down} \AND \isin{#2}{right}}{
            \coordinate (anchor) at ($({#3})!{#5}!({#4})$);
            \node[darkmsgcircle, yshift=-6.0mm] at (anchor) {#6};
            \node[yshift=-2.0mm] at (anchor) {$\rightarrow$};
      }{}
      \ifthenelse{\isin{#1}{up} \AND \isin{#2}{right}}{
            \coordinate (anchor) at ($({#3})!{#5}!({#4})$);
            \node[darkmsgcircle, yshift=6.0mm] at (anchor) {#6};
            \node[yshift=2.0mm] at (anchor) {$\rightarrow$};
      }{}

      \ifthenelse{\isin{#1}{down} \AND \isin{#2}{left}}{
            \coordinate (anchor) at ($({#3})!{#5}!({#4})$);
            \node[darkmsgcircle, yshift=-6.0mm] at (anchor) {#6};
            \node[yshift=-2.0mm] at (anchor) {$\leftarrow$};
      }{}
      \ifthenelse{\isin{#1}{up} \AND \isin{#2}{left}}{
            \coordinate (anchor) at ($({#3})!{#5}!({#4})$);
            \node[darkmsgcircle, yshift=6.0mm] at (anchor) {#6};
            \node[yshift=2.0mm] at (anchor) {$\leftarrow$};
      }{}

      \ifthenelse{\isin{#1}{left} \AND \isin{#2}{up}}{
            \coordinate (anchor) at ($({#3})!{#5}!({#4})$);
            \node[darkmsgcircle, xshift=-5.5mm] at (anchor) {#6};
            \node[xshift=-1.5mm] at (anchor) {$\uparrow$};
      }{}
      \ifthenelse{\isin{#1}{right} \AND \isin{#2}{up}}{
            \coordinate (anchor) at ($({#3})!{#5}!({#4})$);
            \node[darkmsgcircle, xshift=5.5mm] at (anchor) {#6};
            \node[xshift=1.5mm] at (anchor) {$\uparrow$};
      }{}
}

\newcommand{\bwmsg}[6]{
      \ifthenelse{\isin{#1}{left} \AND \isin{#2}{down}}{
            \coordinate (anchor) at ($({#3})!{#5}!({#4})$);
            \node[msgdoublecircle, xshift=-5.5mm] at (anchor) {#6};
            \node[xshift=-1.5mm] at (anchor) {$\downarrow$};
      }{}
      \ifthenelse{\isin{#1}{right} \AND \isin{#2}{down}}{
            \coordinate (anchor) at ($({#3})!{#5}!({#4})$);
            \node[msgdoublecircle, xshift=5.5mm] at (anchor) {#6};
            \node[xshift=1.5mm] at (anchor) {$\downarrow$};
      }{}

      \ifthenelse{\isin{#1}{down} \AND \isin{#2}{right}}{
            \coordinate (anchor) at ($({#3})!{#5}!({#4})$);
            \node[msgdoublecircle, yshift=-6.0mm] at (anchor) {#6};
            \node[yshift=-2.0mm] at (anchor) {$\rightarrow$};
      }{}
      \ifthenelse{\isin{#1}{up} \AND \isin{#2}{right}}{
            \coordinate (anchor) at ($({#3})!{#5}!({#4})$);
            \node[msgdoublecircle, yshift=6.0mm] at (anchor) {#6};
            \node[yshift=2.0mm] at (anchor) {$\rightarrow$};
      }{}

      \ifthenelse{\isin{#1}{down} \AND \isin{#2}{left}}{
            \coordinate (anchor) at ($({#3})!{#5}!({#4})$);
            \node[msgdoublecircle, yshift=-6.0mm] at (anchor) {#6};
            \node[yshift=-2.0mm] at (anchor) {$\leftarrow$};
      }{}
      \ifthenelse{\isin{#1}{up} \AND \isin{#2}{left}}{
            \coordinate (anchor) at ($({#3})!{#5}!({#4})$);
            \node[msgdoublecircle, yshift=6.0mm] at (anchor) {#6};
            \node[yshift=2.0mm] at (anchor) {$\leftarrow$};
      }{}

      \ifthenelse{\isin{#1}{left} \AND \isin{#2}{up}}{
            \coordinate (anchor) at ($({#3})!{#5}!({#4})$);
            \node[msgdoublecircle, xshift=-5.5mm] at (anchor) {#6};
            \node[xshift=-1.5mm] at (anchor) {$\uparrow$};
      }{}
      \ifthenelse{\isin{#1}{right} \AND \isin{#2}{up}}{
            \coordinate (anchor) at ($({#3})!{#5}!({#4})$);
            \node[msgdoublecircle, xshift=5.5mm] at (anchor) {#6};
            \node[xshift=1.5mm] at (anchor) {$\uparrow$};
      }{}
}

\newcommand{\bwdarkmsg}[6]{
      \ifthenelse{\isin{#1}{left} \AND \isin{#2}{down}}{
            \coordinate (anchor) at ($({#3})!{#5}!({#4})$);
            \node[darkmsgdoublecircle, xshift=-5.5mm] at (anchor) {#6};
            \node[xshift=-1.5mm] at (anchor) {$\downarrow$};
      }{}
      \ifthenelse{\isin{#1}{right} \AND \isin{#2}{down}}{
            \coordinate (anchor) at ($({#3})!{#5}!({#4})$);
            \node[darkmsgdoublecircle, xshift=5.5mm] at (anchor) {#6};
            \node[xshift=1.5mm] at (anchor) {$\downarrow$};
      }{}

      \ifthenelse{\isin{#1}{down} \AND \isin{#2}{right}}{
            \coordinate (anchor) at ($({#3})!{#5}!({#4})$);
            \node[darkmsgdoublecircle, yshift=-6.0mm] at (anchor) {#6};
            \node[yshift=-2.0mm] at (anchor) {$\rightarrow$};
      }{}
      \ifthenelse{\isin{#1}{up} \AND \isin{#2}{right}}{
            \coordinate (anchor) at ($({#3})!{#5}!({#4})$);
            \node[darkmsgdoublecircle, yshift=6.0mm] at (anchor) {#6};
            \node[yshift=2.0mm] at (anchor) {$\rightarrow$};
      }{}

      \ifthenelse{\isin{#1}{down} \AND \isin{#2}{left}}{
            \coordinate (anchor) at ($({#3})!{#5}!({#4})$);
            \node[darkmsgdoublecircle, yshift=-6.0mm] at (anchor) {#6};
            \node[yshift=-2.0mm] at (anchor) {$\leftarrow$};
      }{}
      \ifthenelse{\isin{#1}{up} \AND \isin{#2}{left}}{
            \coordinate (anchor) at ($({#3})!{#5}!({#4})$);
            \node[darkmsgdoublecircle, yshift=6.0mm] at (anchor) {#6};
            \node[yshift=2.0mm] at (anchor) {$\leftarrow$};
      }{}

      \ifthenelse{\isin{#1}{left} \AND \isin{#2}{up}}{
            \coordinate (anchor) at ($({#3})!{#5}!({#4})$);
            \node[darkmsgdoublecircle, xshift=-5.5mm] at (anchor) {#6};
            \node[xshift=-1.5mm] at (anchor) {$\uparrow$};
      }{}
      \ifthenelse{\isin{#1}{right} \AND \isin{#2}{up}}{
            \coordinate (anchor) at ($({#3})!{#5}!({#4})$);
            \node[darkmsgdoublecircle, xshift=5.5mm] at (anchor) {#6};
            \node[xshift=1.5mm] at (anchor) {$\uparrow$};
      }{}
}

\usepackage{cite}
\usepackage{array}
\usepackage{graphicx}
\usepackage[hyphens]{url}
\usepackage{todonotes}
\usepackage{fancyhdr}
\usepackage{flushend}
\usepackage{placeins}

\newcommand{\HL}{\mathrm{L}}

\usepackage{amsfonts,amsmath,amssymb}

\renewcommand{\d}[1]{\operatorname{d}\!{#1}}

\newcommand{\N}[1]{\mathcal{N}\!\left({#1}\right)}

\newcommand{\Gam}[1]{\mathcal{G}am\!\left({#1}\right)}
\newcommand{\Ig}[1]{\mathcal{I}g\!\left({#1}\right)}

\begin{document}

\title{A Probabilistic Modeling Approach to Hearing Loss Compensation}
\author{Thijs~van~de~Laar and Bert~de~Vries,~\IEEEmembership{Member,~IEEE}
\thanks{Thijs van de Laar is with the Department of Electrical Engineering, Eindhoven University of Technology, PO Box 513, 5600 MB Eindhoven, the Netherlands (e-mail: t.w.v.d.laar@tue.nl).}
\thanks{Bert de Vries is with GN ReSound, Het Eeuwsel 6, 5612 AS Eindhoven, the Netherlands and with the Department of Electrical Engineering, Eindhoven University of Technology, PO Box 513, 5600 MB Eindhoven, the Netherlands (email: bdevries@ieee.org).}}

\maketitle

\thispagestyle{fancy}

\begin{abstract}
Hearing Aid (HA) algorithms need to be tuned ("fitted") to match the impairment of each specific patient. The lack of a fundamental HA fitting theory is a strong contributing factor to an unsatisfying sound experience for about 20\% of hearing aid patients. This paper proposes a probabilistic modeling approach to the design of HA algorithms. The proposed method relies on a generative probabilistic model for the hearing loss problem and provides for automated inference of the corresponding (1) signal processing algorithm, (2) the fitting solution as well as a principled (3) performance evaluation metric. All three tasks are realized as message passing algorithms in a factor graph representation of the generative model, which in principle allows for fast implementation on hearing aid or mobile device hardware. The methods are theoretically worked out and simulated with a custom-built factor graph toolbox for a specific hearing loss model.     
\end{abstract}


\section{Introduction}
\label{sec:introduction}

\begin{figure}[th!]
        \centering
        \includegraphics[width=8cm]{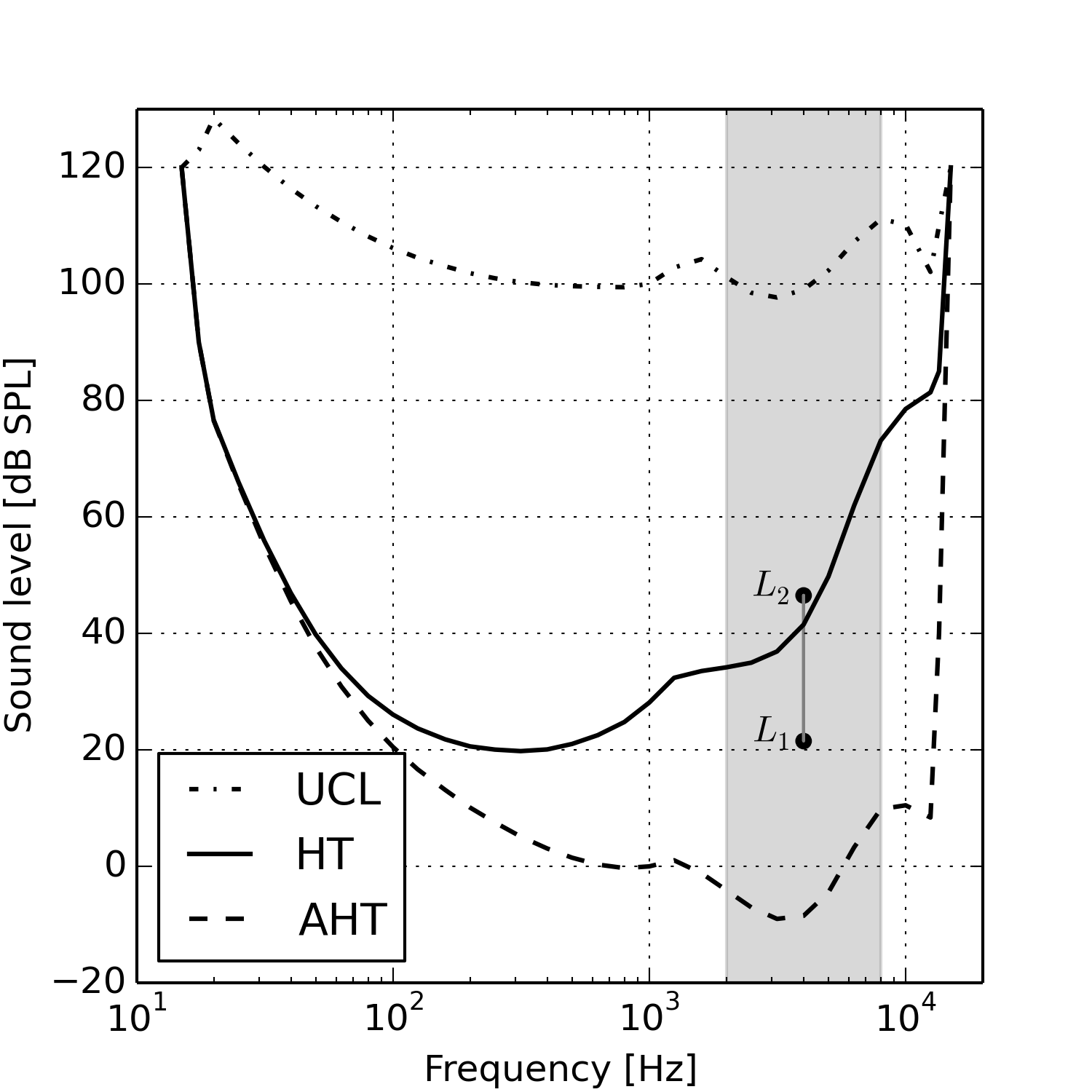}
        \caption{Absolute hearing threshold (AHT) for a normal hearing person (bottom dashed line); a typical hearing threshold for an impaired listener (HT, solid line), and the uncomfortable loudness threshold (UCL, top dashed line), which is usually at similar levels for both a normal and impaired listener. Equal loudness contours for the AHT and the UCL are generated in accordance with the ISO 226:2003 reference standard \cite{iso_226:_2003}, with the AHT at $0$ [phon] and UCL at $100$ [phon].}
\label{fig:equal_loudness}
\end{figure}

\IEEEPARstart{H}{earing} loss is an important problem that affects the quality of life of millions of people. About 15\% of American adults (37.5 million) report problems with hearing \cite{blackwell_summary_2014}.  For most cases, the problem relates to frequency-dependent loss of sensitivity of hearing. In Fig.~\ref{fig:equal_loudness}, the bottom (dashed) curve represents the Absolute Hearing Threshold (AHT) as a function of frequency. The AHT is the just detectable sound level for normal hearing subjects. The top (dash-dotted) curve represents the Uncomfortable Loudness Level (UCL) for the average normal hearing population \cite{iso_226:_2003}. Generally speaking, as we get older, our sensitivity to acoustic inputs deteriorates. In that case, the raised hearing threshold for a particular person may be represented by the middle (solid) curve in Fig.~\ref{fig:equal_loudness}. Now consider an ambient tone at intensity level $L_1$ as indicated by the black circle. This signal would be heard by a normal listener but not by the impaired listener. The primary task of a hearing aid (HA) is to amplify the signal so as to restore normal hearing levels for the ``aided'' impaired listener. Aside from signal processing that compensates for problems that occur due to insertion of the hearing aid itself (e.g., feedback, occlusion, loss of localization), an important challenge in HA signal processing design is to determine the optimal amplification gain $(L_2 - L_1)$.

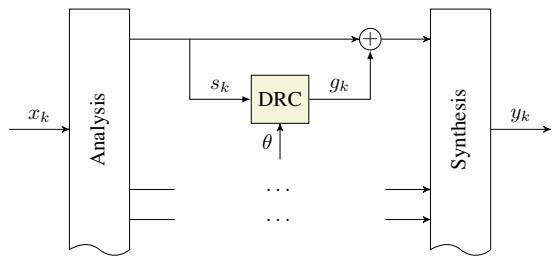
\begin{figure}[ht!]
    \centering
    \scalebox{.8}{
\begin{tikzpicture}
    [node distance=15mm,auto,>=stealth']
    \begin{scope}
        \draw (0,0) -- (0,4) -- (1,4) -- (1,0) to [out=220,in=-40] (0.5,0) to [out=140,in=30] (0,0) -- cycle;
        \node[rotate=90] at (0.5,2.0) {Analysis};
        \draw (6,0) -- (6,4) -- (7,4) -- (7,0) to [out=220,in=-40] (6.5,0) to [out=140,in=30] (6,0) -- cycle;
        \node[rotate=90] at (6.5,2.0) {Synthesis};

        \path[draw] (-1.0,2.0) edge[->] node{$x_k$} (0.0,2.0);
        \path[draw] (7.0,2.0) edge[->] node{$y_k$}(8.0,2.0);

        \node[roundbox] at (5.0,3.5) (plus) {$+$};

        \path[draw] (1.0,3.5) edge[->] (plus);
        \path[draw] (plus) edge[->] (6.0,3.5);

        \node[box, fill=beige] at (3.5, 2.5) (drc) {DRC};
        \path[line] (2.0,3.5) -- (2.0,2.5) -> node{$s_k$} (drc);
        \path[line] (drc) -- node{$g_k$} (5.0,2.5) -> (plus);

        \path[draw] (3.5, 1.5) edge[->] node{$\theta$} (drc);

        \node at (3.5, 1.0) {$\dots$};
        \node at (3.5, 0.5) {$\dots$};
        \path[draw] (1.0, 1.0) -- (1.75, 1.0);
        \path[draw] (5.25, 1.0) edge[->] (6.0, 1.0);
        \path[draw] (1.0, 0.5) -- (1.75, 0.5);
        \path[draw] (5.25, 0.5) edge[->] (6.0, 0.5);
    \end{scope}
\end{tikzpicture}
    }
    \caption{A typical HA signal processing scheme with dynamic range compression (DRC) gain agents in the frequency bands of a filter bank. $x_k$ and $y_k$ represent HA input and output audio signals respectively, $s_k$ is the estimated log-power of the input signal (in dB SPL) in a frequency band at time step $k$ and $g_k$ is the hearing loss compensation gain (in dB). $\theta$ represents the DRC tuning parameters, which generally include a compression ratio, compression threshold and attack- and release-times. Usually, the DRC circuits in each frequency band operate independently from each other.}
    \label{fig:filter_bank}
\end{figure}

\begin{figure}[hb!]
        \centering
        \includegraphics[width=8cm]{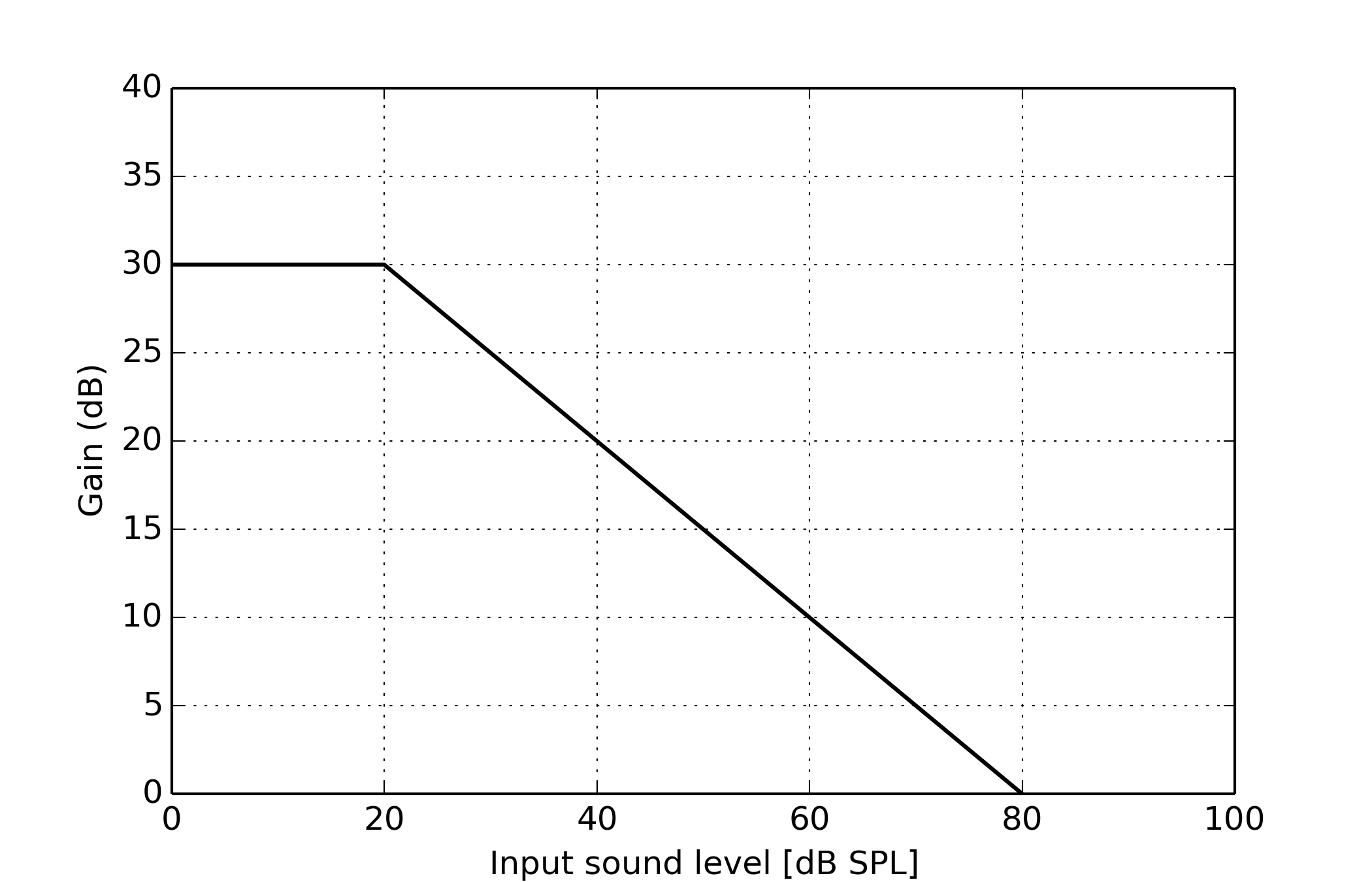}
        \caption{A typical gain vs.\ signal intensity relation in one frequency band for a hearing aid dynamic range compressor.}
\label{fig:gain_curve}
\end{figure}

Technically, the optimal gain depends on the specific hearing loss of the user and turns out to be both frequency and intensity-level dependent. In commercial hearing aids, amplification is generally based on (multi-channel) dynamic range compression (DRC) processing in the frequency bands of a filter bank, see Fig.~\ref{fig:filter_bank}. A typical gain vs.\ signal intensity relation in one frequency band of a DRC circuit is shown in Fig.~\ref{fig:gain_curve}. The gain is maximal for low input levels and remains constant with growing input levels until a Compression Threshold (CT), after which the gain decreases linearly (on a logarithmic scale). The slope of the gain decrease is determined by the compression ratio $\mathrm{CR} \triangleq \Delta \text{input} / \Delta (\text{input}+\text{gain})$, which is a characteristic parameter for DRC algorithms. Aside from CT and CR, a DRC circuit is typically also parametrized by attack and release time constants (AT and RT, respectively) to control the dynamic behavior. The crucial problem of estimating good values for the parameters CT, CR, AT and RT is called the \emph{fitting} problem.

Modern hearing aid fitting strategies set compression ratios by prescriptive rules, e.g., the NAL \cite{byrne_nal-nl1_2001} and DSL \cite{cornelisse_input/output_1995} rules are very widely used. For the dynamic parameters (AT and RT) no standard fitting rules exist and most HA manufacturers offer slight variations on known dynamic recipes such as slow-acting ('automatic volume control') and fact-acting ('syllabic') compression \cite{moore_choice_2008}. The goal of tuning HA settings through prescriptive fitting rules is to provide a decent 'first-fit' since any personal preferences of HA patients are not taken into account. Usually, a fine-tuning session with a professional audiologist is needed optimize the HA fit to a specific patient. 

Currently, a hearing aid designer can design a satisfactory hearing aid for a given patient in any given situation. However, the designer does not have the advance information to design a solution for an unknown client in an unknown, volatile acoustic world. As a result, while hearing aids do improve the lives of many people, there is still a fraction of about 20\% of patients who are not satisfied with the sound quality of their hearing aids \cite{kochkin_marketrak_2010}.  

In order to increase HA patient satisfaction levels, hearing aid design should be more \emph{personalizable} by the patient himself. Usable HA personalization needs to balance information gain with patient effort. While we need more preference feedback from patients to fine-tune their HAs, we do not want to substantially increase the cognitive burden-of-elicitation on HA patients. Hence there is a need to develop personalized hearing aid design methods that make optimal use of sparsely available preference data. In this paper we present a novel methodology to design \emph{personalized} hearing aid algorithms. Four main questions naturally arise:

\begin{itemize}
\item How to describe HA design as a personalized design process?
\item Which signal processing circuit should be used by the hearing aid device? We call this the Signal Processing (\textbf{SP}) task.
\item How to estimate the tuning parameters of the signal processing algorithm? This is the Parameter Estimation (\textbf{PE}) task (a.k.a. fitting task in hearing aid parlance).
\item How to evaluate the performance of the signal processing algorithm relative to an alternative? This we call the Model Comparison (\textbf{MC}) task.
\end{itemize}

We address all four issues. In order to optimally utilize the information present in preference data, we use a probabilistic modelling approach. In our framework, we first specify the uncertainty about the exact hearing loss problem by a probabilistic model. Next we use a data base of preferred HA input-output pairs (the `training' database) to personalize (i.e., refine) the problem statement. The solutions to the SP, PE and MC tasks will be automatically \emph{inferred} though optimal Bayesian reasoning. We summarize our contributions here:

\subsubsection*{Personalized Design Process}
\begin{itemize}
\item We describe a personalized HA design process in Sec.~\ref{sec:personalized_hearing_aid_design}. Our design method suggests in-situ alternative HA settings in case a patient is unsatisfied with the current performance of his HA. Additionally, the proposed method collects a data base of preferred audio processing examples. This data base can be used to personalize the SP, PE and MC tasks. 
\end{itemize}

\subsubsection*{Signal processing}
\begin{itemize}
\item We describe a probabilistic model for hearing loss compensation and use Bayesian probabilistic reasoning to automatically infer the signal processing solution (Secs.~\ref{sec:gen_model_spec} and \ref{sec:SP_methods}).
\item We demonstrate that the resulting circuit is a dynamic range compressor (DRC) with adaptive time constants, (Secs.~\ref{sec:SP_results} and \ref{sec:SP_discussion}). Crucially, this DRC circuit is inferred rather than designed.
\item In contrast to existing literature, the proposed solution method includes an explicit description of the hearing loss problem that it solves. As a result, our design method provides an explicit entry point for the engineer to describe the problem (Sec.~\ref{sec:SP_discussion}). 
\end{itemize}

\subsubsection*{Parameter estimation}

\begin{itemize}
\item Parameter estimation (fitting) is also described as a probabilistic inference problem (Sec.~\ref{sec:PE_methods}).
\item Since optimal parameters are inferred, there is no need for prescriptive fitting rules. Instead, fitting prescriptions are implicitly specified by patient evaluations during listening trials.  
\item Whereas existing prescriptive rules only fit a small subset of HA algorithm parameters, the proposed method in principle fits all HA parameters. 
\end{itemize}

\subsubsection*{Model comparison}
\begin{itemize}
\item In Sec.~\ref{sec:MC_methods}, we describe a HA algorithm performance metric (the Bayes factor) as a probabilistic inference task. Specifically, we describe how to compare so-called nested HA algorithm structures.
\item In contrast to existing popular HA performance measures, the Bayes Factor is a data-driven performance metric, which makes it personalizable.  
\end{itemize}

\subsubsection*{Realization}
\begin{itemize}
\item We describe how the SP, PE and MC tasks can be uniformly realized by an efficient framework for probabilistic reasoning, based on message passing in a Forney-style Factor Graph (Sec.~\ref{sec:ffg}). This framework allows for realization on ultra low-power DSP circuits that are used in hearing aids. 
\end{itemize}

Moreover, in our framework, the SP, PE and MC tasks are based on the same reasoning process and support each other in a consistent way, e.g., inference of the signal processing algorithm turns out to be a crucial part of the parameter estimation process. Our focus will be on presenting the fundamental methods for automated HA algorithm design, rather than presenting a specific improved HA algorithm. Therefore, a clinical evaluation is not included in this work.

\section{The Solution Framework}
\label{sec:solution_framework}

\subsection{In-situ personalized hearing aid design}
\label{sec:personalized_hearing_aid_design}
This section aims to describe a method for in-situ collection of preferred HA processing examples. Under in-situ conditions, each positive patient evaluation leads to an additional example of preferred HA processing data. The accumulated preference data set is used to personalize the SP, PE and MC tasks. In order to get the HA system into preferred modes, we explore new HA settings each time after the patient has expressed dissatisfaction with the ongoing HA processing.
     
\begin{figure}[t]
        \centering
        \includegraphics[width=8cm]{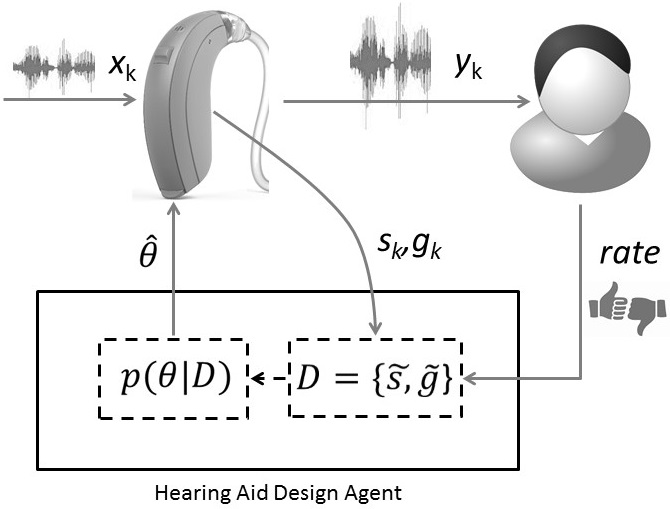}
        \caption{Flow diagram of our proposed HA design method based on in-situ preference feedback. The Hearing Aid Design Agent (HADA) responds to negative user feedback by sending alternative tuning parameter values ($\hat \theta$) to the HA. In response to positive user feedback, HADA updates its state-of knowledge about preferred HA parameter settings, which is represented by the distribution $p(\theta|D)$.}
\label{fig:PHLC-architecture}
\end{figure}

Consider the scenario as depicted in Fig.~\ref{fig:PHLC-architecture}. The system under study comprises a hearing aid device, a HA patient and a Hearing Aid Design Agent (HADA). The object of study in this paper is the HADA, which designs a signal processing module for the hearing aid device on the basis of preference feedback from the patient. In principle, the signal processing module can be any HA module, e.g., a noise reduction or feedback cancellation module. In this paper we focus on the HADA as a Hearing Loss Compensation (HLC) module. The input and output at time $k$ of the HLC module are (log-power) $s_k$ and (log-gain) $g_k$, respectively, and we assume that $\theta$ is a vector of tuning parameters for the HLC module. The HLC module connects to other modules in the HA, for instance to a filter bank module and transducers. Ultimately, the HA device produces an audio output signal $y_k$ (see also Fig.\ref{fig:filter_bank}) for the HA patient. We assume that the patient can signal a binary appraisal of HA performance to the HADA. When the patient is dissatisfied with the HA performance, HADA is expected to respond by sending alternative parameter values $\hat \theta$ to the HA. Thus, from the HA patient's perspective, a new HA algorithm is applied whenever he submits a negative response. The patient then listens to a differently processed sound and he may decide to signal a new appraisal. We label the time spent on a specific parameter setting as a \emph{trial}. In this framework, HA design is an always-on incremental process of consecutive trials.  

An important goal for HADA is that the alternative parameter values $\hat \theta$ are \emph{interesting} to the HA patient. The issue of selecting interesting actions has been studied widely in the reinforcement learning literature as the exploitation-exploration issue \cite{berger-tal_exploration-exploitation_2014,botvinick_model-based_2014}. Our approach is based on maintaining a probability distribution $p(\theta|D)$ over the tuning parameters, where $D$ is observed data. This probability should be interpreted as a (normalized) \emph{preference function} for the tuning parameters, i.e., if $p(\theta_1|D)>p(\theta_2|D)$ then in our model the HA patient prefers $\theta_1$ over $\theta_2$. New trials (e.g., after the HA patient submits a negative appraisal) are generated by drawing a sample from the preference function:

\begin{equation}
	\hat \theta \sim p(\theta|D)\,.
\end{equation}

This strategy for selecting interesting trials is also known as \emph{Thompson sampling} and balances the exploitation-exploration trade-off \cite{chapelle_empirical_2011}.  

If the HA patient is content with the current HA output signal, he may choose to signal a positive appraisal to the HADA. In this case, HADA stores a few seconds of input and output signals ($\tilde s$ and $\tilde g$, where the overhead-tilde notation indicates sequences of indeterminate length) of the signal processing module to the training data base $D$. Over time, as the user submits multiple positive responses, HADA accumulates a data base of preferred input-output pairs. In Sec.~\ref{sec:PE_methods} we discuss how HADA uses the data base of preferred input-output pairs to learn the preference distribution $p(\theta|D)$ for the HA user.

In summary, we propose a principled in-situ design method for HA algorithms that uses binary performance feedback from patients. The method aims to minimize the burden-of-interaction on the end user. In this framework, HA design is an always-on incremental process that learns the unobserved hearing loss of an unknown patient in an unknown and volatile acoustic environment.

\subsection{Automated inference for signal processing, fitting and performance evaluation}
\label{sec:inference_for_sp_pe_mc}

In order to personalize the SP, PE and MC tasks, we pose the hearing aid algorithm design problem as a problem of probabilistic inference. This approach starts with the specification of a so-called generative probabilistic model. This paper focuses on a signal processing algorithm for hearing loss compensation. The signal processing algorithm applies a compensation gain $g_k$ to an input log-power level $s_k$ at each time step $k$. We denote the specific choice of algorithm architecture (i.e., the set of equations) by a model selection parameter $m$. A model is controlled by a model-specific set of tuning parameters $\theta$. Note that the explicit specification of the model choice $m$ and model parameters $\theta$ is a task for the algorithm design engineer. A \emph{generative probabilistic model} specifies a joint probability distribution over all relevant variables in the system under study. In our case, the probabilistic generative model becomes 

\begin{equation}
\label{eq:generative_model-1}
p(g, s, \theta, m)\,,
\end{equation}

\noindent which relates the gain sequence $g = (g_0, g_1, \ldots)$ at all time steps to the log-power levels $s = (s_1, s_2, \ldots)$ (both in dB).  

The tasks of interest, namely signal processing, parameter estimation and model comparison, can be specified as inference problems on the generative model.  

Firstly, in the setting of this paper, signal processing consists of (recursively) estimating the compensation gain $g_k$ at time step $k$ after observing power levels $s^k \triangleq (s_1,\ldots,s_k)$. In other words, the signal processing (SP) task at time step $k$ is described by computing 

\begin{equation}
\label{eq:sp}
p(g_k|s^k,\theta,m) = \frac{\idotsint p(s^k,g^k,\theta,m)\d{g_0}\dots\d{g_{k-1}}}{\idotsint p(s^k,g^k,\theta, m)\d{g_0}\dots\d{g_k}}\,.
\end{equation} 

Secondly, parameter estimation relates to estimating good parameter values $\theta$. The parameters are estimated from a \emph{training set} of signal processing examples. Assume that a training data set of preferred input-output pairs $D \triangleq \{ (\tilde s, \tilde g) \}$ has been collected by the patient, e.g., by the  in-situ collection method as described in  Sec.~\ref{sec:personalized_hearing_aid_design}. The parameter estimation (PE) problem then consists of solving the following inference problem:

\begin{equation}
\label{eq:pe}
p(\theta|\tilde s,\tilde g,m) = \frac{p(\tilde s,\tilde g,\theta,m)}{\int p(\tilde s,\tilde g,\theta,m)\d{\theta}} \,. 
\end{equation}

Thirdly, the model comparison task expresses the (relative) performance of model $m$ versus an alternative model $m^\dagger$ as a hearing aid signal processing algorithm. Here, we take a Bayesian model comparison viewpoint and compute the posterior probability ratio for model $m$ versus an alternative model $m^\dagger$ with fitted parameters $\theta^\dagger$:

\begin{equation}
\label{eq:mc}
\frac{p(m|\tilde s,\tilde g)}{p(m^\dagger|\tilde s,\tilde g)} = \frac{\int p(\tilde s,\tilde g,\theta,m) \d{\theta}}{\int p(\tilde s,\tilde g,\theta^\dagger,m^\dagger) \d{\theta^\dagger}}\,. 
\end{equation}

Crucially, both the PE task (which determines $\theta$) and MC task (which determines $m$) are personalized since they depend on a personal collection of preferred audio processing examples. As a consequence, the signal processing algorithm is also personalized, since signal processing depends both on the algorithm choice ($m$) and the algorithm fit ($\theta$), cf. Eq.~\ref{eq:sp}.

In summary, the problem statement for a probabilistic modelling approach to HA algorithm design comprises three main issues: 

\begin{enumerate}
\renewcommand{\labelenumi}{\textbf{\theenumi}}
\renewcommand{\theenumi}{S.\arabic{enumi}}
\item How to specify the generative model Eq.~\ref{eq:generative_model-1}? (see Sec.~\ref{sec:gen_model_spec} for answer.)
\item How to acquire an efficient training data set (answered in Sec.~\ref{sec:personalized_hearing_aid_design}.)
\item How to execute the inference tasks in Eqs.~\ref{eq:sp}, \,\ref{eq:pe} and \ref{eq:mc} (see Secs.~\ref{sec:ffg}, \ref{sec:SP_methods}, \ref{sec:PE_methods} and \ref{sec:MC_methods} for answers.) 
\end{enumerate}
 
The most interesting aspect of the proposed approach is that all relevant tasks (SP, PE and MC) can be automatically implemented by executing an inference task on the generative model. These inference tasks mostly involve computation of large integrals over a subset of the variables in the generative model Eq.~\ref{eq:generative_model-1}. In Sec.~\ref{sec:ffg} we will introduce message passing in factor graphs as an efficient framework for computing the required integrals.

The rest of the paper focuses on a technical solution approach to issues \textbf{S.1} and \textbf{S.3}. We will work out issues  \textbf{S.1} and \textbf{S.3} in detail for a hearing loss compensation module. For the purpose of this paper, we assume that a training dataset (issue \textbf{S.2}) has been selected by the user.

\section{Methods}

\subsection{Generative model specification}
\label{sec:gen_model_spec}

In this section, we detail the specification of our generative model $p(s,g,\theta, m)$. Since we only discuss one system architecture in this section, the model selection variable $m$ is temporarily dropped from the equations. As can be seen from Eqs.~\ref{eq:sp}, \ref{eq:pe} and \ref{eq:mc}, hearing aid signal processing and design mostly involves integrating over subsets of variables. In order to render computation of these integrals tractable, we factorize the generative model at time step $n$ into a set of smaller models \cite{koller_probabilistic_2009} by
 
\begin{align}\label{eq:full_generative_model}
p(&s^n,g^n,\theta) \notag\\
&= p(g_0)\, p(\theta)\, \prod_{k=1}^n p(s_k|g_k,\theta) \, p(g_k|g_{k-1},\theta) \,.
\end{align}

The generative model factorization breaks the generative model into a Markov chain, i.e., the value of any variable at time step $k$ is determined by the value of system variables at step $k-1$ together with new observations at step $k$. We distinguish an \emph{observation model} $p(s_k|g_k,\theta)$ and a \emph{gain transition model} $p(g_k|g_{k-1},\theta)$. In Eq.~\ref{eq:full_generative_model}, $p(g_0)$ and $p(\theta)$ are prior distributions that are usually chosen to be vague (e.g., with large variance). Next, we work out the details for the observation and gain transition models and further detail the composition of model parameters $\theta$. 

\subsubsection{Observation model}
\label{sec:observation_model}

The observation model $p(s_k|g_k,\theta)$ specifies that the observed power input at step $k$ depends (only) on the compensation gain at the same time step. We use this model to specify our goal to restore loudness for the impaired listener. We use a simple but generally accepted model for hearing loss as proposed by Zurek \cite{zurek_hearing_2007}. Zurek's hearing loss model maps received sound levels (levels that enter the ear) to attenuated sound levels for a hearing impaired listener (in each frequency band). In our problem statement the received sound level is the input level with added compensation gain. Therefore, we model the hearing loss as a function of the original input sound level $s_k$ plus compensation gain $g_k$ in [dB HL] through

\begin{align}
\label{eq:Zurek_model}
    \HL(&s_k+g_k; \alpha, \beta) \notag\\
    &= \left\{ 
        \begin{array}{c l}
            {0} & {\mbox{if } s_k+g_k < -\beta/\alpha}\\
            {\alpha (s_k+g_k) + \beta} & {\mbox{if } -\beta/\alpha \leq s_k+g_k < -\beta/(\alpha-1)}\\
            {s_k+g_k} & {\mbox{otherwise.}} 
        \end{array}
    \right.  
\end{align}

\begin{figure}[hb!]
    \centering
    \includegraphics[width=8cm]{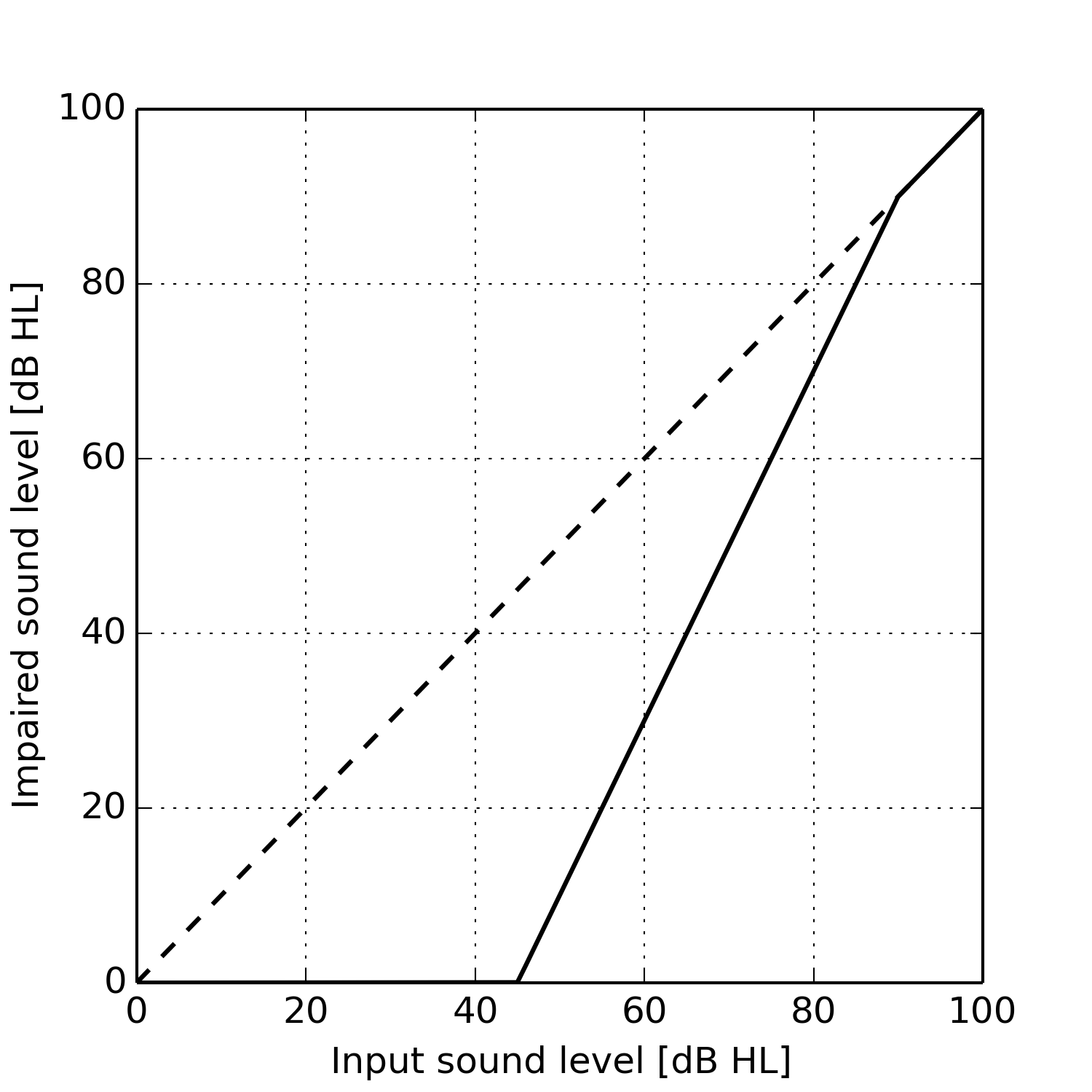}
    \caption{Zurek's hearing loss model $\HL$ (solid line) with parameters $\alpha = 2, \beta = -90$, which corresponds to a hearing threshold at 45 [dB HL] and recruitment threshold at 90 [dB HL].}
    \label{fig:hl_model}
\end{figure}

Zurek's model contains tuning parameters $\alpha$ and $\beta$ that need to be tuned (fitted) for each frequency band for each listener in order to obtain the preferred signal processing settings. Fig.~\ref{fig:hl_model} shows a typical hearing loss curve for Zurek's model. Note that the model contains two breaking points. We recognize the abscissa of the first breaking point $\mathrm{HT} \triangleq -\beta/\alpha$ as the impaired listener's \emph{hearing threshold}, and the abscissa of the second breaking point $\mathrm{RT} \triangleq -\beta/(\alpha-1)$ as the \emph{recruitment threshold}. Zurek's model includes hearing loss (for inputs less than $\mathrm{HT}$), fast loudness recruitment (for inputs between $\mathrm{HT}$ and $\mathrm{RT}$) and normal loudness perception for loud inputs (above $\mathrm{RT}$).

The primary goal of hearing loss compensation is to amplify input signals such that sound levels for an aided impaired-listener approximate normal sound levels. Formally, we model the idea of ``approximately'' through adding zero-mean Gaussian noise with variance $\vartheta$. This variance regulates the strictness with which the compensation constraint is enforced. The observation model then evaluates to 

\begin{equation}
    p(s_k | g_k, \alpha, \beta, \vartheta) \propto \N{s_k|\HL(s_k+g_k; \alpha, \beta), \vartheta} \label{eq:observation_model}\,,
\end{equation}

\noindent where we have substituted $\theta$ with three model parameters, namely $\alpha$, $\beta$ and $\vartheta$. We are free to extend this set, and will add one more parameter when specifying the gain transition model.

Note that the variable $s_k$ in Eq.~\ref{eq:observation_model} appears in the `mean' parameter for the normal distribution and therefore Eq.~\ref{eq:observation_model} does not represent a Gaussian model for $s_k$. Still, Eq.~\ref{eq:observation_model} represents a proper constraint on the signal processing solution and Eq.~\ref{eq:full_generative_model} specifies a valid generative model factorization.

\subsubsection{Gain transition model}
\label{sec:gain_transition}

The generative model requires a specification of the gain transition model $p(g_k|g_{k-1},\theta)$. We want to penalize fast gain changes since they may lead to distortions in the output audio signal. This is modeled by a Gaussian random walk model for the gain transitions as specified by Eq.~\ref{eq:gain_change_prior}. Since allowed gain changes are usually small, we choose to parametrize the model with a precision parameter $\gamma$ that governs how strictly we penalize gain changes:

\begin{equation}
\label{eq:gain_change_prior}
    p(g_k|g_{k-1},\gamma) = \N{g_k | g_{k-1}, \gamma^{-1}}\,.
\end{equation}

\noindent At this point we have specified the complete set of tuning parameters $\theta = \{\alpha, \beta, \vartheta, \gamma\}$. 

In summary, the set of equations Eqs.~\ref{eq:full_generative_model}, \ref{eq:Zurek_model}, \ref{eq:observation_model} and \ref{eq:gain_change_prior} fully specify the generative model for hearing loss compensation. In this model, the input power levels $s_k$ are considered observed variables, since the analysis branch of the filter bank provides their values. The inference of the gain $g_k$ is the primary objective of the signal processing algorithm. In this model, the gain is an unobserved (hidden) variable.

In principle, the tuning parameters are also unobserved and it is the objective of the parameter estimation task to infer adequate values. As discussed, in a real HA algorithm, the generative HLC model is applied independently for each frequency band in the filter bank.

More elaborate gain transition models can also be incorporated (or left out altogether). For modeling (acoustic) context-dependent gain transitions for example, the Gaussian noise sources in the gain transition model could be extended to the more general hierarchical Gaussian filter \cite{mathys_hierarchical_2012}. Similarly, we could have chosen a more sophisticated model for hearing impairment, such as described in \cite{jepsen_characterizing_2011}. In the end, the question which model assumptions are best is answered by a model comparison analysis (as discussed in Sec.~\ref{sec:MC_methods}). 

\subsection{Forney-style factor graphs}
\label{sec:ffg}

In Sec.~\ref{sec:inference_for_sp_pe_mc}, we showed that the SP, PE and MC tasks involve computation of complicated integrals on the generative model. A popular generic framework for Bayesian integration is based on Monte Carlo sampling. However, sampling methods suffer from a high degree of computational complexity \cite{mackay_information_2003}. Here, we are specifically interested in computationally efficient methods and the framework of Probabilistic Graphical Models fits this profile well \cite{bishop_pattern_2006}, \cite{koller_probabilistic_2009}. In particular for signal processing applications, the \emph{Forney-style Factor Graph} (FFG) framework has shown to provide an efficient framework for solving inference problems \cite{kschischang_factor_2001}, \cite{loeliger_factor_2007}. In this section, we discuss the FFG framework for efficiently computing marginalization integrals in probabilistic models.  

In order to render Bayesian integrals tractable, it is important to impose a high degree of factorization on the generative model. As an illustrative example, consider the (factorized) probabilistic model 

\begin{equation}
    p(x_1, \dots, x_5) = f_a(x_1, x_2) \cdot f_b(x_2,x_3,x_5) \cdot f_c(x_4, x_5) \,. \label{eq:factorization}
\end{equation}

\begin{figure}[ht!]
\centering
\scalebox{.8}{
\begin{tikzpicture}
    [node distance=20mm,auto,>=stealth']
    \begin{scope}

        \node (fl) {};
        \node[box, right of=fl, node distance=15mm] (fa) {$f_a$};
        \node[box, right of=fa] (fb) {$f_b$};
        \node[box, right of=fb, node distance=25mm] (fc) {$f_c$};
        \node[right of=fc, node distance=15mm] (fr) {};
        \node[below of=fb, node distance=15mm] (fd) {};

        \path[line] (fl) edge[->] node[anchor=north]{$x_1$} (fa);
        \path[line] (fa) edge[->] node[anchor=north]{$x_2$} (fb);
        \path[line] (fb) edge[->] node[anchor=north]{$x_5$} (fc);
        \path[line] (fc) edge[->] node[anchor=north]{$x_4$} (fr);
        \path[line] (fb) edge[<-] node[anchor=west]{$x_3$} (fd);

        \msg{up}{right}{fa}{fb}{0.5}{1}
        \msg{up}{right}{fb}{fc}{0.38}{2}
        \msg{up}{left}{fb}{fc}{0.62}{3}

        \draw[dashed] (-0.1,0.7) rectangle (2.1,-0.7);
        \draw[dashed] (-0.35,1.0) rectangle (4.1,-1.6);
        \draw[dashed] (5.4,0.7) rectangle (7.6,-0.7);
    \end{scope}    
\end{tikzpicture}
}
\caption{A Forney-style factor graph representation of Eq.~\ref{eq:inference_example}. }
\label{fig:simple_ffg}
\end{figure}
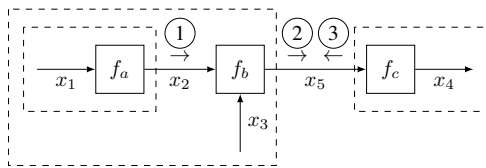

The factorization of Eq.~\ref{eq:factorization} is drawn as a graph in Fig.~\ref{fig:simple_ffg}, where the nodes and edges correspond to the factors and variables respectively. Note that each factor connects only to edges whose corresponding variables are in the factor's arguments. This graphical representation of a factorized function is called a Forney-style Factor Graph (FFG), \cite{forney_codes_2001}. It is common to indicate an \emph{observed} variable by a smaller closed square (e.g., see $s_k$ in Fig.~\ref{fig:generative_drc}).    

Now assume that we want to obtain the marginal probability distribution over $x_5$:

\begin{equation}
    p(x_5) = \idotsint p(x_1, \dots, x_5) \d{x_1} \dots \d{x_4} \label{eq:marginal}\,.
\end{equation}

Substituting Eq.~\ref{eq:factorization} in Eq.~\ref{eq:marginal} and rearranging the integrals according to the distributive law, yields

\begin{align}
    p(x_5) = &\underbrace{\iint\!f_b(x_2, x_3, x_5) \d{x_3} \overbrace{\int\!f_a(x_1, x_2) \d{x_1}}^{\circled{1}}\d{x_2}}_{\circled{2}} \notag\\
    &\times \underbrace{\int\!f_c(x_4, x_5) \d{x_4}}_{\circled{3}} \,.\label{eq:inference_example}
\end{align}

\noindent The large integral of Eq.~\ref{eq:marginal} breaks into a set of smaller integrals, denoted by the circled numbers. In the factor graph of Fig.~\ref{fig:simple_ffg}, the sub-integrals summarize the probability distribution over the box they leave. Therefore, the procedure of computing a sub-integral is also referred to as \emph{closing the box}. The summaries are called \emph{messages}, and computing these nested integrals allows for stepwise marginalization, referred to as \emph{message passing}. The particular marginalization process described here is known as the sum-product algorithm \cite{gallager_low-density_1962,loeliger_introduction_2004}. Note that messages flow in both directions over an edge. In Fig.~\ref{fig:simple_ffg}, messages $\circled{2}$ and \circled{3} respectively summarize information about $x_5$ by the left and right parts (relative to edge $x_5$) of the graph. Although FFGs are principally undirected graphs, the edges are often drawn with arrowheads in order to  distinguish the notation for a forward message $\overrightarrow{\mu}(x_5)$ (message $\circled{2}$) from a backward message  $\overleftarrow{\mu}(x_5)$ (message \circled{3}). The marginal distribution over a variable (e.g., $p(x_5)$) is obtained by multiplying colliding messages ($\circled{2}$ and $\circled{3}$). 

\subsubsection{Branching Points and Equality nodes}
\label{sec:equality_nodes}

\begin{figure}[th!]
    \centering
    \scalebox{.9}{
\begin{tikzpicture}
    [node distance=15mm,auto,>=stealth']
    \begin{scope}

        \node[box] (eq) {$=$};
        \node[left of=eq] (l) {$\dots$};
        \node[right of=eq] (r) {$\dots$};
        \node[below of=eq] (b) {$\dots$};

        \path[line] (l) edge[->] node{$x$} node[anchor=north, pos=0.15]{$_{\overrightarrow{\mu}_{x}(x)}$} (eq);
        \path[line] (b) edge[->] node{$y$} node[anchor=west]{$_{\overrightarrow{\mu}_{y}(y)}$} (eq);
        \path[line] (r) edge[<-] node[anchor=south]{$z$} node[anchor=north, pos=0.10]{$_{\overrightarrow{\mu}_{z}(z)}$} (eq);
    \end{scope}    
\end{tikzpicture}
    }
    \caption{An equality node constrains the marginals over the variables $x$, $y$ and $z$ to be equal through the factor $f(x, y, z) = \delta(z-x)\,\delta(z-y)$.}
    \label{fig:eq_node}
\end{figure}
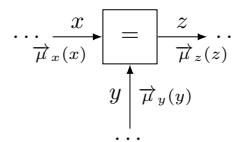

In FFGs, an edge represents a variable and connects to factors that contain that variable in their argument set. This implies that an FFG can only be constructed for a factorized distribution where each variable name appears maximally in two factors. We can circumvent this limitation by introducing factors that copy variables (and update the factorization with the new variable). These factors are usually called \emph{equality nodes} and represent branching points in the graph, see Fig.~\ref{fig:eq_node}. The equality constraint node implements the factor $f(x, y, z) = \delta(z-x)\,\delta(z-y)$. 

For illustrative purposes, let us compute the sum-product update rule for outgoing message $\overrightarrow{\mu}_{z}(z)$ as a function of incoming messages $\overrightarrow{\mu}_{x}(x)$ and $ \overrightarrow{\mu}_{y}(y)$ for the equality node  (see also \cite{loeliger_introduction_2004, korl_factor_2005}). The sum-product rule prescribes that the outgoing message is computed by first multiplying all incoming messages with the factor function, followed by integrating (summing) over all incoming variables. In other words, the sum-product rule for the equality constraint node evaluates to 

\begin{align}
    \label{eq:eq_node_update}
    \overrightarrow{\mu}_{\!z}(z) &= \iint \overrightarrow{\mu}_{x}(x) \overrightarrow{\mu}_{y}(y) f(x, y, z) \d{x} \d{y} \notag\\
    &= \iint \overrightarrow{\mu}_{x}(x) \overrightarrow{\mu}_{y}(y) \delta(z-x) \delta(z-y) \d{x} \d{y} \notag \\
    &= \overrightarrow{\mu}_{x}(z) \overrightarrow{\mu}_{y}(z)\,.
\end{align}

The concept of equality nodes makes it possible to draw a Forney-style factor graph for any factorized probability distribution. Note that the equality node intuitively implements Bayes rule, as it fuses information from two sources $x$ and $y$ into $z$. In practical applications, $\overrightarrow{\mu}_{x}(z)$ often represents prior information about $z$ that gets updated by a likelihood function $\overrightarrow{\mu}_{y}(z)$ (or vice versa).

Message passing on FFGs facilitates efficient computation of marginals (integrals) on the probabilistic model that the graph represents. For many elementary node functions, the sum-product message update rules can be analytically solved and stored in a look-up table. In this case, inference on probabilistic models amounts to sequentially executing the appropriate message update rules. Message passing in FFGs thus provides a framework for efficient automated inference in factorized probability distributions. For a more extensive introduction to FFGs we refer the reader to the excellent tutorials in \cite{loeliger_introduction_2004, loeliger_factor_2007}.

\subsubsection{Factor graph for the generative model}

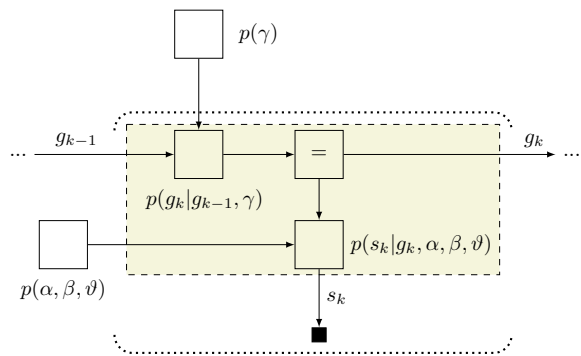
\begin{figure}[ht!]
    \centering
    \scalebox{0.8}{
\begin{tikzpicture}
    [node distance=15mm,auto,>=stealth']
    \begin{scope}

        \node (dots) {$...$};
        \node[box, right of=dots, node distance=3.0cm] (p_gk) {};
        \node[below of=p_gk, node distance=0.75cm, xshift=1mm] () {$p(g_k|g_{k-1}, \gamma)$};
        \node[box, above of=p_gk, node distance=20mm] (tgk) {};
        \node[right of=tgk, node distance=1.0cm] () {$p(\gamma)$};
        \node[box, right of=p_gk, node distance=2.0cm] (eqk) {$=$};
        \node[box, below of=eqk] (p_sk) {};
        \node[right of=p_sk, node distance=1.7cm] () {$p(s_k|g_k,\alpha, \beta, \vartheta)$};
        \node[box, left of=p_sk, node distance=4.25cm] (tsk) {};
        \node[below of=tsk, node distance=0.75cm] () {$p(\alpha, \beta, \vartheta)$};
        \node[clamped, below of=p_sk] (sk) {};

        \node[right of=eqk, node distance=4.2cm] (rdots) {$...$};

        \path[line] (dots) edge[->] node[pos=0.3]{$g_{k-1}$} (p_gk);
        \path[line] (tgk) edge[->] (p_gk);
        \path[line] (p_gk) edge[->] (eqk);
        \path[line] (eqk) edge[->] (p_sk);
        \path[line] (tsk) edge[->] (p_sk);
        \path[line] (p_sk) edge[->] node{$s_k$} (sk);
        
        \path[line] (eqk) edge[->, pos=0.9] node{$g_k$} (rdots);

        \brackets{1.9}{7.9}{0.4}{-3.0}

        \begin{pgfonlayer}{bg}
            \draw[dashed, fill=beige] (1.8,0.5) rectangle (8.0,-2.0);
        \end{pgfonlayer}
    \end{scope}    
\end{tikzpicture}
    }
    \caption{A factor graph representation for the generative hearing loss compensation model Eq.~\ref{eq:full_generative_model}. The shaded area balances loudness restoration with gain evolution. Edges are principally undirected; arrowheads merely anchor message directions. We follow some notational conventions from \cite{reller_state-space_2012}. Dotted arcs indicate a repetition of the enveloped section. Solid (small) black nodes denote that the corresponding edge (variable) is observed.}
    \label{fig:generative_drc}
\end{figure}

We can now draw a factor graph for the generative hearing loss compensation model as specified by the factorization in Eq.~\ref{eq:full_generative_model}., see Fig.\ref{fig:generative_drc}. This graph shows the factorization for one time step in one frequency band.  
The highlighted factors trade off (through an equality constraint node) the simultaneous goals of loudness restoration and (reduction of) signal distortion. 

\subsection{Signal processing}
\label{sec:SP_methods}

As discussed in Sec.~\ref{sec:inference_for_sp_pe_mc}, signal processing relates to computing $p(g_k|s^k,\theta,m)$ for each time step $k$. During execution of the signal processing task, we keep the parameters $\theta$ and model choice $m$ fixed. Dropping $\theta$ and $m$ from the equations, note that

\begin{align}
    \label{eq:sp_recursion}
    p(g_k&|s^k) \propto \underbrace{p(g_k, s^k)}_{\text{posterior}} \notag\\
    &= \int p(s_k, g_k|g_{k-1}) \, p(g_{k-1}, s^{k-1}) \d{g_{k-1}} \notag \\
    &= \underbrace{p(s_k|g_k)}_{\text{observation}} \, \int \underbrace{p(g_k | g_{k-1})}_{\text{transition}} \, \underbrace{p(g_{k-1}, s^{k-1})}_{\text{prior}} \d{g_{k-1}}\,.
\end{align}

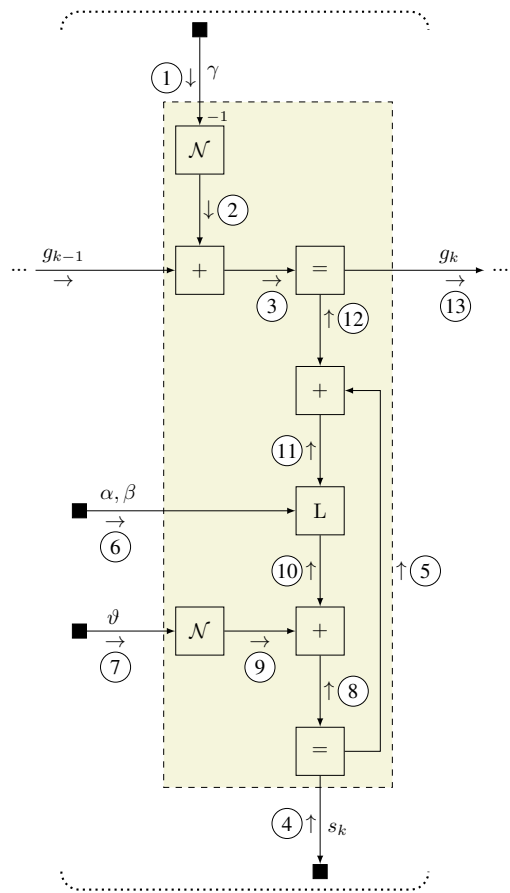
\begin{figure}
    \centering
    \scalebox{.8}{
\begin{tikzpicture}
    [node distance=20mm,auto,>=stealth']
    \begin{scope}

        \node[] (g_n_min_1) {$...$};
        \node[box, right of=g_n_min_1, node distance=30mm] (g_plus) {$+$};
        \node[box, right of=g_plus] (g_eq) {$=$};
        \node[right of=g_eq, node distance=30mm] (g_n) {$...$};
        \node[box, above of=g_plus] (n_top) {$\mathcal{N}$};
        \node[yshift=5mm, xshift=3mm] at (n_top) {$^{-1}$};
        \node[clamped, above of=n_top] (w) {}; 
        \node[box, below of=g_eq] (s_plus) {$+$};
        \node[box, below of=s_plus] (L) {$\mathrm{L}$};
        \node[box, below of=L] (v_add) {$+$};
        \node[box, below of=v_add] (s_eq) {$=$};
        \node[clamped, below of=s_eq] (s_n_bottom) {};
        \node[box, left of=v_add] (n_bottom) {$\mathcal{N}$};
        \node[clamped, left of=n_bottom] (v) {};
        \node[clamped, left of=L, node distance=40mm] (theta_ab) {};

        \path[line] (w) edge[->] node[pos=0.4]{$\gamma$} (n_top);
        \path[line] (n_top) edge[->] (g_plus);
        \path[line] (g_n_min_1) edge[->] node[anchor=north, pos=0.2]{$\rightarrow$} node[pos=0.2]{$g_{k-1}$} (g_plus);
        \path[line] (g_plus) edge[->] (g_eq);
        \path[line] (g_eq) edge[->] node[pos=0.75]{$g_k$} (g_n);
        \path[line] (g_eq) edge[->] (s_plus);
        \path[line] (s_plus) -> (L);
        \path[line] (v_add) edge[->] (s_eq); 
        \path[line] (s_eq) edge[->] node[pos=0.6]{$s_k$} (s_n_bottom);
        \path[line] (theta_ab) edge[->] node[anchor=south, pos=0.15]{$\alpha, \beta$} (L);
        \path[line] (L) edge[->] (v_add);
        \path[line] (v_add) edge[<-] (n_bottom);
        \path[line] (n_bottom) edge[<-] node[anchor=south, pos=0.7]{$\vartheta$} (v);
        \draw[line] (s_eq) -- ($(s_eq)+(1.0,0)$) -- ($(s_plus)+(1.0,0)$) -> (s_plus);

        \coordinate (s_eq_right) at ($(s_eq)+(1.2,0)$);
        \coordinate (s_plus_right) at ($(s_plus)+(1.2,0)$);

        \msg{left}{down}{w}{n_top}{0.4}{1}
        \msg{right}{down}{n_top}{g_plus}{0.5}{2}
        \msg{down}{right}{g_plus}{g_eq}{0.6}{3}
        \msg{left}{up}{s_eq}{s_n_bottom}{0.6}{4}
        \msg{right}{up}{s_eq_right}{s_plus_right}{0.5}{5}
        \msg{down}{right}{theta_ab}{L}{0.15}{6}
        \msg{down}{right}{n_bottom}{v}{0.7}{7}
        \msg{right}{up}{v_add}{s_eq}{0.5}{8}
        \msg{down}{right}{v_add}{n_bottom}{0.5}{9}
        \msg{left}{up}{L}{v_add}{0.5}{10}
        \msg{left}{up}{s_plus}{L}{0.5}{11}
        \msg{right}{up}{g_eq}{s_plus}{0.4}{12}
        \msg{down}{right}{g_eq}{g_n}{0.75}{13}

        \brackets{1.0}{6.5}{4.0}{-10.0}

        \begin{pgfonlayer}{bg}
            \draw[dashed, fill=beige] ($(n_top)+(-0.6,0.8)$) rectangle ($(s_eq)+(1.2,-0.6)$);
        \end{pgfonlayer}
    \end{scope}    
\end{tikzpicture}
    }
    \caption{FFG and message passing schedule for signal processing. The shaded area balances loudness restoration with signal distortion.}
    \label{fig:sp-filtering}
\end{figure}

As a result, we conclude that signal processing  can be executed through \emph{recursive} inference, where the posterior $p(g_k,s^k)$ depends on the prior $p(g_{k-1}, s^{k-1})$, the gain transition model $p(g_k | g_{k-1})$ and the observation model $p(s_k|g_k)$. This inference task for step $k$ can be executed by message passing in a Forney-style factor graph. In Fig.~\ref{fig:sp-filtering}, the generative model for step $k$ and the corresponding message passing schedule for signal processing is depicted. This FFG is constructed from the generative model by substituting the specific gain transition and observation models and fixing (indicated by smaller solid black boxes) the values of the observed input $s_k$ and parameters $\theta = \{\alpha, \beta, \vartheta, \gamma\}$. Hence, in thirteen updates (messages) per time step, this factor graph implementation automatically derives the signal processing task.   

\subsection{Parameter estimation}
\label{sec:PE_methods}

In Sec.~\ref{sec:inference_for_sp_pe_mc}, we discussed that the parameter estimation problem corresponds to the inference task $p(\theta|D,m)$. Next, we show that parameter estimation can be executed by a message passing algorithm on the factor graph for the generative model.

We start by introducing a time dependence in $\theta$ by defining $\theta^n \triangleq (\theta_0,\theta_1, \ldots, \theta_n)$ and a \emph{parameter transition model} $p(\theta_k|\theta_{k-1})$. 

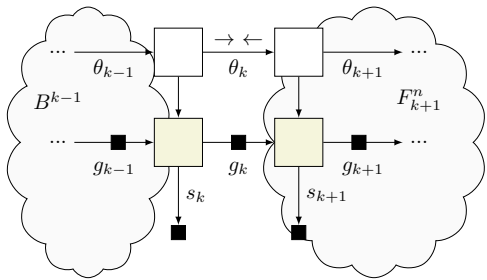
\begin{figure}
    \centering
    \scalebox{0.8}{
\begin{tikzpicture}
    [node distance=20mm,auto,>=stealth']

    \node[] (theta_dots_l) {$...$}; 
    \node[box, right of=theta_dots_l, fill=white] (theta_l) {}; 
    \node[box, right of=theta_l, fill=white] (theta_r) {}; 
    \node[right of=theta_r] (theta_dots_r) {$...$}; 

    \node[below of=theta_dots_l, node distance=15mm] (g_dots_l) {$...$}; 
    \node[box, right of=g_dots_l, fill=beige] (g_l) {}; 
    \node[box, right of=g_l, fill=beige] (g_r) {}; 
    \node[right of=g_r] (g_dots_r) {$...$}; 

    \node[clamped, right of=g_dots_l, node distance=10mm] (g_k_min) {};
    \node[clamped, right of=g_k_min] (g_k) {}; 
    \node[clamped, right of=g_k] (g_k_plus) {}; 

    \node[clamped, below of=g_l, node distance=15mm] (s_k) {};
    \node[clamped, below of=g_r, node distance=15mm] (s_k_plus) {}; 

    \path[line] (theta_dots_l) edge[->] node[anchor=north]{$\theta_{k-1}$} (theta_l);
    \path[line] (theta_l) edge[->] node[anchor=north]{$\theta_{k}$} node[anchor=south, pos=0.3]{$\rightarrow$} node[anchor=south, pos=0.7]{$\leftarrow$} (theta_r);
    \path[line] (theta_r) edge[->] node[anchor=north]{$\theta_{k+1}$} (theta_dots_r);

    \path[line] (g_dots_l) edge[->] node[anchor=north, yshift=-2mm]{$g_{k-1}$} (g_l);
    \path[line] (g_l) edge[->] node[anchor=north, yshift=-2mm]{$g_{k}$} (g_r);
    \path[line] (g_r) edge[->] node[anchor=north, yshift=-2mm]{$g_{k+1}$} (g_dots_r);

    \path[line] (theta_l) edge[->] (g_l);
    \path[line] (theta_r) edge[->] (g_r);

    \path[line] (g_l) edge[->] node[]{$s_{k}$} (s_k);
    \path[line] (g_r) edge[->] node[]{$s_{k+1}$} (s_k_plus);

    \begin{pgfonlayer}{bg}
		\node[cloud, draw,cloud puffs=14,cloud puff arc=120, aspect=2, inner ysep=1em, minimum height=4.5cm, minimum width=2.75cm, fill=lightgrey] at (0.5,-1.5) {};
		\node[] at (0,-0.80) {$B^{k-1}$};
		\node[cloud, draw,cloud puffs=14,cloud puff arc=120, aspect=2, inner ysep=1em, minimum height=4.5cm, minimum width=3.75cm, fill=lightgrey] at (5.25,-1.5) {};
		\node[] at (6.0,-0.80) {$F_{k+1}^n$};
    \end{pgfonlayer}

\end{tikzpicture}
    }
    \caption{Overview FFG for parameter estimation, with input levels and gains observed. The grey nodes correspond with the grey shaded area of the generative model (Fig.~\ref{fig:generative_drc}).}
    \label{fig:fw_bw_derivation}
\end{figure}

Following \cite{minka_hidden_1999}, we focus on the local factorization around time point $k$. For simplicity, we assume a \emph{given} training data set $\{s^n,g^n\}$ of input-output examples of length $n$. Let us group the variables before and after $k<n$ into $B^{k-1} = \{s^{k-1},g^{k-1},\theta^{k-1}\}$ and $F_{k+1}^n = \{s_{k+1}^n,g_{k+1}^n,\theta_{k+1}^n\}$ respectively, where the combined sub-superscript indicates a range, e.g., $s_{k+1}^n = (s_{k+1}, \dots, s_n)$ (see also Fig.~\ref{fig:fw_bw_derivation}).

The posterior for $\theta_k$, based on sequences $\{s^n,g^n\}$ with $n>k$ can be evaluated to the product of a \emph{forward} and \emph{backward message}: 

\begin{align}
    p(&\theta_k|s^n, g^n) \propto p(s^n, g^n, \theta_k) \notag\\
   &= \idotsint p(s^n,g^n,\theta^n) \d{\theta_0} \dots \d{\theta_{k-1}} \d{\theta_{k+1}} \dots \d{\theta_n} \notag\\
   &= \idotsint p(B^{k-1},g_k,\theta_k) \, p(s_k|g_k,\theta_k) \notag \\
        &\quad \times p(F_{k+1}^n|g_k,\theta_k) \d{\theta_0} \dots \d{\theta_{k-1}} \d{\theta_{k+1}} \dots \d{\theta_n} \notag\\
    &= \left( \idotsint p(B^{k-1},g_k,\theta_k) \d{\theta_0} \dots \d{\theta_{k-1}} \right) \, p(s_k|\theta_k,g_k) \notag \\
        &\quad  \times  \left( \idotsint p(F_{k+1}^n|g_k,\theta_k) \d{\theta_{k+1}} \dots \d{\theta_n} \right) \notag \\
    &= p(s^{k-1},g^k,\theta_k) \, p(s_k|g_k,\theta_k) \, p(s_{k+1}^n, g_{k+1}^n|g_k,\theta_k) \notag\\
    &= \underbrace{p(s^k,g^k,\theta_k)}_{\text{forward msg}} \, \underbrace{p(s_{k+1}^n, g_{k+1}^n|g_k,\theta_k)}_{\text{backward msg}}\,. \label{eq:fw_bw}
\end{align}

The forward and backward messages of Eq.~\ref{eq:fw_bw} can be calculated by forward and backward recursions (respectively) over time. Substituting the factorization of the generative model into the forward message of Eq.\ref{eq:fw_bw} leads to the (forward) recursion

\begin{align}
   p(s^k,&g^k,\theta_k) = \underbrace{p(s_k|g_k,\theta_k)}_{\text{observation}} \, \underbrace{p(g_k|g_{k-1},\theta_k)}_{\text{gain transition}} \notag \\
        & \times \int \underbrace{p(\theta_k|\theta_{k-1})}_{\text{par. transition}}\, \underbrace{p(s^{k-1},g^{k-1},\theta_{k-1})}_{\text{prior}} \d{\theta_{k-1}}\,. \label{eq:fw}
\end{align}

Note the resemblance of Eq.~\ref{eq:fw} to the signal processing recursion Eq.~\ref{eq:sp_recursion}. Similarly, the backward recursion follows from the second term of Eq.\ref{eq:fw_bw}:

\begin{align}
    p(s_k^n, &g_k^n|g_{k-1},\theta_{k-1}) = \int  \underbrace{p(s_k|g_k,\theta_k)}_{\text{observation}} \, \underbrace{p(g_k|g_{k-1},\theta_k)}_{\text{gain transition}} \notag\\
        &\times \underbrace{p(\theta_k|\theta_{k-1})}_{\text{par. transition}} \, \underbrace{p(s_{k+1}^n, g_{k+1}^n|g_k,\theta_k)}_{\text{prior}} \d{\theta_k}\,. \label{eq:bw}
\end{align}

Parameter estimation can thus be executed by recursive inference of Eqs.~\ref{eq:fw} and \ref{eq:bw}, followed by Eq.~\ref{eq:fw_bw}. This procedure is also known as the \emph{forward-backward algorithm} \cite{rabiner_tutorial_1989}. 

\begin{figure*}[ht!]
    \centering
    \scalebox{.8}{
\begin{tikzpicture}
    [node distance=20mm,auto,>=stealth']

    \begin{scope}
        \node[clamped, node distance=20mm] (g_n_min_1) {};
        \node[box, right of=g_n_min_1, node distance=20mm] (g_plus) {$+$};
        \node[box, right of=g_plus] (g_eq) {$=$};
        \node[clamped, right of=g_eq] (g_n) {};
        \node[box, above of=g_plus] (n_top) {$\mathcal{N}$};
        \node[yshift=5mm, xshift=3mm] at (n_top) {$^{-1}$};
        \node[box, above of=n_top] (w) {$=$}; 
        \node[left of=w, node distance=45mm] (w_l) {$...$}; 
        \node[right of=w, node distance=55mm] (w_r) {$...$}; 
        \node[box, below of=g_eq] (s_plus) {$+$};
        \node[box, below of=s_plus] (L) {$\mathrm{L}$};
        \node[box, below of=L] (v_add) {$+$};
        \node[box, below of=v_add] (s_eq) {$=$};
        \node[clamped, below of=s_eq] (s_k) {};
        \node[box, left of=v_add] (n_bottom) {$\mathcal{N}$};
        \node[box] (v_eq) at (0.5,-10.5) {$=$};
        \node[box] (ab_eq) at (-0.5,-11.5) {$=$};
        \node[left of=v_eq, node distance=30mm] (v_l) {$...$}; 
        \node[right of=v_eq, node distance=70mm] (v_r) {$...$}; 
        \node[left of=ab_eq, node distance=20mm] (ab_l) {$...$}; 
        \node[right of=ab_eq, node distance=80mm] (ab_r) {$...$}; 

        \path[line] (v_l) edge[->, pos=0.1] node[anchor=north]{$\rightarrow$} node{$\vartheta'$} (v_eq);
        \path[line] (v_eq) edge[->] node[pos=0.9]{$\vartheta''$} node[anchor=north, pos=0.9]{$\leftarrow$} (v_r);
        \path[line] (ab_l) edge[->] node[anchor=north, pos=0.2]{$\rightarrow$} node[pos=0.1]{$\alpha', \beta'$} (ab_eq);
        \path[line] (ab_eq) edge[->] node[pos=0.95]{$\alpha'', \beta''$} node[anchor=north, pos=0.915]{$\leftarrow$} (ab_r);
        \path[line] (w) edge[->] node[anchor=east]{$\gamma$} (n_top);
        \path[line] (w_l) edge[->] node[anchor=north, pos=0.1]{$\rightarrow$} node[pos=0.1]{$\gamma'$} (w);
        \path[line] (w) edge[->] node[pos=0.9]{$\gamma''$} node[anchor=north, pos=0.9]{$\leftarrow$} (w_r);
        \path[line] (n_top) edge[->] (g_plus);
        \path[line] (g_n_min_1) edge[->] node[pos=0.4]{$g_{k-1}$} (g_plus);
        \path[line] (g_plus) edge[->] (g_eq);
        \path[line] (g_eq) edge[->] node[pos=0.75]{$g_k$} (g_n);
        \path[line] (g_eq) edge[->] (s_plus);
        \path[line] (s_plus) -> (L);
        \path[line] (s_k) edge[<-] node[anchor=west, pos=0.4]{$s_k$} (s_eq);
        \draw[line] (s_eq) -- ($(s_eq)+(1.0,0)$) -- ($(s_plus)+(1.0,0)$) -> (s_plus);
        \path[line] (L) edge[->] (v_add);
        \path[line] (v_add) edge[<-] (n_bottom);
        \path[line] (v_add) edge[->] (s_eq);
        \path[line] (ab_eq) -- node[pos=0.375]{$\alpha, \beta$} ($(ab_eq)+(0,7.5)$) -> (L);
        \path[line] (v_eq) -- node[pos=0.4]{$\vartheta$} ($(v_eq)+(0,4.5)$) -> (n_bottom);

        \coordinate (ab_up) at ($(ab_eq)+(0,7.5)$);
        \coordinate (v_up) at ($(v_eq)+(0,4.5)$);
        \coordinate (s_eq_right) at ($(s_eq)+(1.2,0)$);
        \coordinate (s_plus_right) at ($(s_plus)+(1.2,0)$);

        \brackets{-1.0}{6.0}{4.5}{-12}

        \begin{pgfonlayer}{bg}
            \draw[dashed, fill=beige] ($(n_top)+(-0.8,0.8)$) rectangle ($(s_eq)+(1.2,-0.6)$);
            \draw[dotted, line width = 1pt] ($(n_bottom)+(-0.6,2.6)$) rectangle ($(v_add)+(0.6,-0.6)$);
        \end{pgfonlayer}

        \msg{down}{right}{g_n_min_1}{g_plus}{0.4}{1}
        \msg{left}{down}{n_top}{g_plus}{0.4}{\underline{2}} 
        \msg{down}{right}{g_plus}{g_eq}{0.5}{3} 
        \msg{down}{left}{g_eq}{g_n}{0.75}{4}
        \msg{left}{down}{g_eq}{s_plus}{0.5}{5} 
        \msg{left}{up}{s_k}{s_eq}{0.4}{6}
        \msg{right}{up}{s_eq_right}{s_plus_right}{0.5}{7}
        \msg{left}{down}{s_plus}{L}{0.5}{8} 
        \msg{right}{up}{s_plus}{L}{0.5}{\underline{9}}
        \msg{right}{up}{g_eq}{s_plus}{0.5}{10}
        \msg{up}{left}{g_plus}{g_eq}{0.5}{11}
        \msg{right}{up}{n_top}{g_plus}{0.4}{12}
        \msg{left}{up}{v_add}{s_eq}{0.6}{13} 
        \msg{right}{down}{v_add}{s_eq}{0.6}{\underline{14}}
        \msg{right}{up}{w}{n_top}{0.35}{\underline{15}}
        \msg{right}{down}{v_eq}{v_up}{0.7}{\underline{16}}
        \msg{right}{down}{ab_eq}{ab_up}{0.875}{\underline{17}}
    \end{scope}

    \begin{scope}[xshift=11.5cm]
        \node[clamped, node distance=20mm] (g_n_min_1) {};
        \node[box, right of=g_n_min_1, node distance=20mm] (g_plus) {$+$};
        \node[box, right of=g_plus] (g_eq) {$=$};
        \node[clamped, right of=g_eq] (g_n) {};
        \node[box, above of=g_plus] (n_top) {$\mathcal{N}$};
        \node[yshift=5mm, xshift=3mm] at (n_top) {$^{-1}$};
        \node[box, above of=n_top] (w) {$=$}; 
        \node[left of=w, node distance=45mm] (w_l) {$...$}; 
        \node[right of=w, node distance=55mm] (w_r) {$...$}; 
        \node[box, below of=g_eq] (s_plus) {$+$};
        \node[box, below of=s_plus] (L) {$\mathrm{L}$};
        \node[box, below of=L] (v_add) {$+$};
        \node[box, below of=v_add] (s_eq) {$=$};
        \node[clamped, below of=s_eq] (s_k) {};
        \node[box, left of=v_add] (n_bottom) {$\mathcal{N}$};
        \node[box] (v_eq) at (0.5,-10.5) {$=$};
        \node[box] (ab_eq) at (-0.5,-11.5) {$=$};
        \node[left of=v_eq, node distance=30mm] (v_l) {$...$}; 
        \node[right of=v_eq, node distance=70mm] (v_r) {$...$}; 
        \node[left of=ab_eq, node distance=20mm] (ab_l) {$...$}; 
        \node[right of=ab_eq, node distance=80mm] (ab_r) {$...$}; 

        \path[line] (v_l) edge[->, pos=0.1] node[anchor=north]{$\rightarrow$} node{$\vartheta'$} (v_eq);
        \path[line] (v_eq) edge[->] node[pos=0.9]{$\vartheta''$} node[anchor=north, pos=0.9]{$\leftarrow$} (v_r);
        \path[line] (ab_l) edge[->] node[anchor=north, pos=0.2]{$\rightarrow$} node[pos=0.1]{$\alpha', \beta'$} (ab_eq);
        \path[line] (ab_eq) edge[->] node[pos=0.95]{$\alpha'', \beta''$} node[anchor=north, pos=0.915]{$\leftarrow$} (ab_r);
        \path[line] (w) edge[->] node[anchor=west]{$\gamma$} (n_top);
        \path[line] (w_l) edge[->] node[anchor=north, pos=0.1]{$\rightarrow$} node[pos=0.1]{$\gamma'$} (w);
        \path[line] (w) edge[->] node[pos=0.9]{$\gamma''$} node[anchor=north, pos=0.9]{$\leftarrow$} (w_r);
        \path[line] (n_top) edge[->] (g_plus);
        \path[line] (g_n_min_1) edge[->] node[pos=0.4]{$g_{k-1}$} (g_plus);
        \path[line] (g_plus) edge[->] (g_eq);
        \path[line] (g_eq) edge[->] node[pos=0.75]{$g_k$} (g_n);
        \path[line] (g_eq) edge[->] (s_plus);
        \path[line] (s_plus) -> (L);
        \path[line] (s_k) edge[<-] node[anchor=west, pos=0.4]{$s_k$} (s_eq);
        \draw[line] (s_eq) -- ($(s_eq)+(1.0,0)$) -- ($(s_plus)+(1.0,0)$) -> (s_plus);
        \path[line] (L) edge[->] (v_add);
        \path[line] (v_add) edge[<-] (n_bottom);
        \path[line] (v_add) edge[->] (s_eq);
        \path[line] (ab_eq) -- node[pos=0.375]{$\alpha, \beta$} ($(ab_eq)+(0,7.5)$) -> (L);
        \path[line] (v_eq) -- node[pos=0.4]{$\vartheta$} ($(v_eq)+(0,4.5)$) -> (n_bottom);

        \coordinate (ab_up) at ($(ab_eq)+(0,7.5)$);
        \coordinate (v_up) at ($(v_eq)+(0,4.5)$);
        \coordinate (s_eq_right) at ($(s_eq)+(1.2,0)$);
        \coordinate (s_plus_right) at ($(s_plus)+(1.2,0)$);

        \brackets{-1.0}{6.0}{4.5}{-12}

        \begin{pgfonlayer}{bg}
            \draw[dashed, fill=beige] ($(n_top)+(-0.8,0.8)$) rectangle ($(s_eq)+(1.2,-0.6)$);
            \draw[dotted, line width = 1pt] ($(n_bottom)+(-0.6,2.6)$) rectangle ($(v_add)+(0.6,-0.6)$);
        \end{pgfonlayer}

        \msg{down}{right}{w}{w_r}{0.5}{18}
        \darkmsg{down}{left}{w_l}{w}{0.5}{19}
        \darkmsg{left}{down}{w}{n_top}{0.35}{20}
        \msg{up}{right}{v_eq}{v_r}{0.2857}{21}
        \darkmsg{up}{left}{v_l}{v_eq}{0.4}{22}
        \darkmsg{left}{up}{v_eq}{v_up}{0.7}{23}
        \msg{up}{right}{ab_eq}{ab_r}{0.375}{24}
        \darkmsg{up}{left}{ab_l}{ab_eq}{0.6}{25}
        \darkmsg{left}{up}{ab_eq}{ab_up}{0.875}{26}
    \end{scope}
\end{tikzpicture}
    }
    \caption{FFG and variational update scheme for PE. The dotted (inner) box signifies that enclosed nodes are merged into a single node function. The schedule on the left computes messages $\protect\smallcircled{\underline{15}}$, $\protect\smallcircled{\underline{16}}$ and $\protect\smallcircled{\underline{17}}$ towards the parameters from the messages coming from the observations ($\protect\smallcircled{1}$, $\protect\smallcircled{4}$ and $\protect\smallcircled{6}$). The right schedule mixes beliefs between time slices. White messages correspond with the forward and black messages with the backward pass. Upon execution, the left schedule is alternated with the right schedule.}
    \label{fig:fitting_vmp}
\end{figure*}
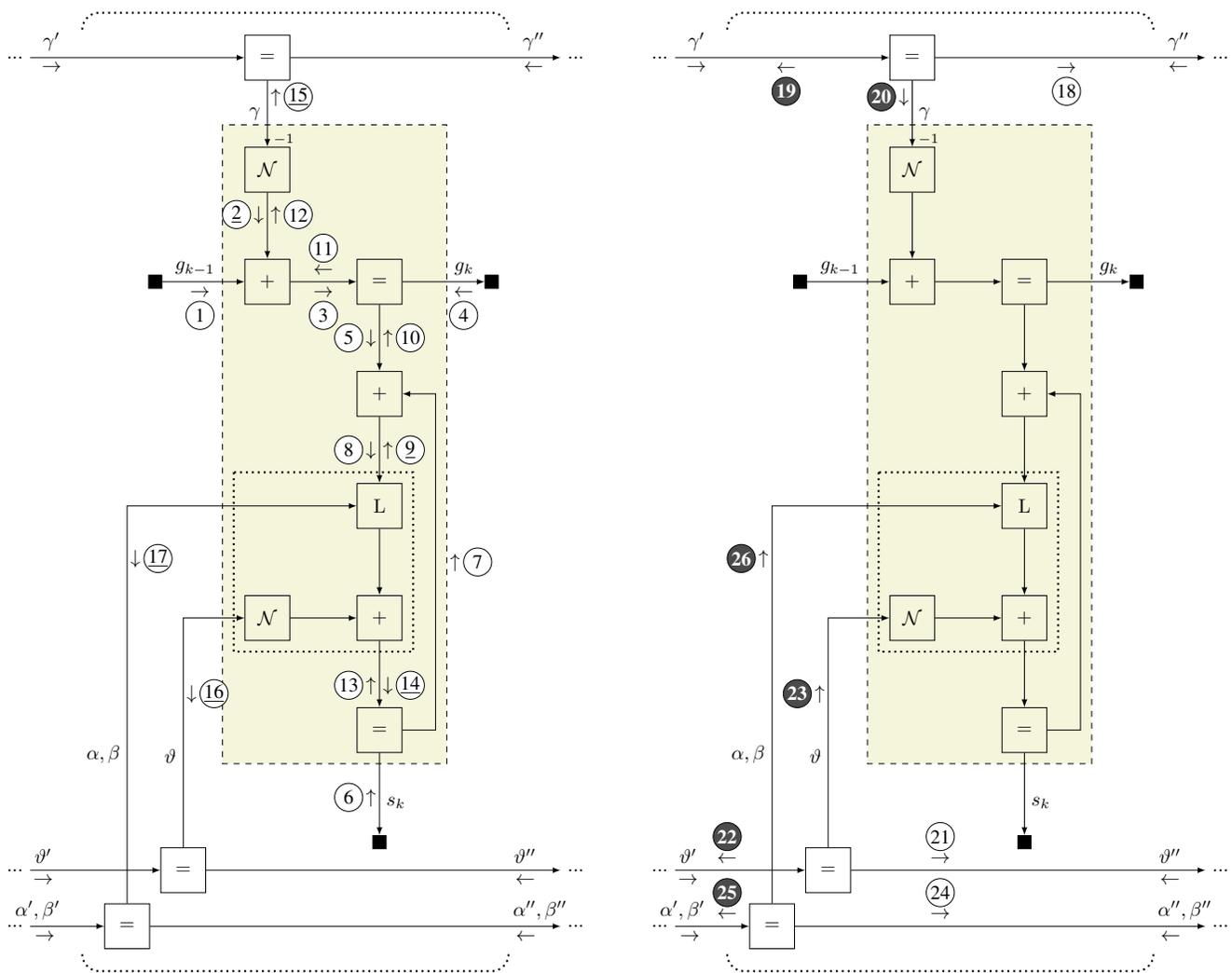

Similarly to the signal processing task, this recursive procedure can be implemented by message passing on a factor graph of the generative model. The (sum-product) message update rules for parameter estimation yield messages that are not always a member of the Gaussian family. This is a complication that would lead to intractability issues in message representation in the rest of the graph \cite{korl_factor_2005}. In order to remain within the Gaussian family, and thus keep computations tractable, we applied an approximate inference scheme called (mean-field) Variational Message Passing (VMP). In the mean-field variant of VMP, the true posterior $p(\theta |s^n, g^n)$ is approximated by a factorized distribution $q(\theta) = q(\alpha)\,q(\beta)\,q(\vartheta)\,q(\gamma)$. A detailed discussion of VMP lies outside the scope of this paper. We followed the implementation for VMP in Forney-style factor graphs as described by Dauwels \cite{dauwels_variational_2007}, to which we also refer for more details on the specific message update rules. Fig.~\ref{fig:fitting_vmp} shows one time step for both the forward and backward passes. In Fig.~\ref{fig:fitting_vmp}, underlined messages are calculated through the variational update rule instead of the sum-product update rule. 

The message passing schedule works both for estimating (slowly) time-varying and (unknown) fixed parameters. Setting the parameter transition model to 

\begin{equation}
    \label{eq:par-transition-is-delta}
    p(\theta_k|\theta_{k-1})=\delta(\theta_k-\theta_{k-1})\,,
\end{equation}

\noindent yields $p(\theta|s^n, g^n) = p(\theta_k|s^n, g^n)$ for any $k$ and consequently results in a single parameter estimate for the whole training sequence. Other choices for the parameter transition model, e.g., a Gaussian random walk $p(\theta_k|\theta_{k-1}) = \mathcal{N}(\theta_k | \theta_{k-1},\vartheta_\theta)$, treat the parameters as (slowly varying) state variables.

\subsection{Model comparison}
\label{sec:MC_methods}

Up to this point, all inference is based around the assumption that the algorithm choice (the set of equations, indicated by $m$) is fixed. However, it is of interest to explore alternative algorithm choices as well. Following the Bayesian framework, we measure the performance of model $m$ versus an alternative model $m^\dagger$ by the posterior probability ratio $p(m^\dagger|D) / p(m|D)$, where $D$ is an (arbitrary) training set. Note that we use the terms `model' and `algorithm' interchangeably here; the HA community tends to speak about `algorithms' whereas the (Bayesian) machine learning community prefers the more generic term `model'. The posterior model ratio,

\begin{equation}
    \frac{p(m^\dagger|D)}{p(m|D)} = \underbrace{\frac{p(D|m^\dagger)}{p(D|m)}}_{\text{Bayes Factor}} \, \frac{p(m^\dagger)}{p(m)}\,,  \label{eq:BF_intro}
\end{equation}

\noindent is proportional to the model likelihood ratio, which is also called the \emph{Bayes Factor} (BF). In the machine learning literature it is common to focus model comparison treatments on computing the (data-dependent) Bayes factor, since the choice for model priors (right term) usually reflects subjective knowledge and intentions. All else being equal, the Bayes factor naturally prefers simple over complicated models by penalizing the inclusion of extra parameters \cite{raftery_bayesian_1995}. 

Crucially, the Bayes Factor provides an \emph{objective} yet \emph{personalized} performance metric for HA algorithms. It is objective in the sense that it is an analytical expression that by itself does not depend on specific (hearing) domain knowledge. Yet it is a personalized measure of performance since it depends on the patient-specific data base of preferred audio processing examples ($D$). 

It often makes sense to limit the set of models under investigation to a so-called ``nested'' set. A model $m^{\dagger}$ is nested in a model $m$ if some parameters $\phi$ of $m$ can be constrained to yield $m^\dagger$. The Bayes factor (in favor of the alternative model) can then be derived through the encompassing prior approach \cite{klugkist_bayes_2007}, and is expressed as

\begin{equation}
    \mathrm{BF} = \frac{\int p(\phi|D, m) I_{\mathcal{O}}(\phi)\d{\phi}}{\int p(\phi|m) I_{\mathcal{O}}(\phi)\d{\phi}} \label{eq:ep_ratio}\,,
\end{equation}

\noindent where $I_{\mathcal{O}}(\phi)$ enforces the nesting constraint, defined as

\begin{equation} \label{eq:nesting_constraint}
    I_{\mathcal{O}}(\phi) = \left\{
        \begin{array}{c l}
            {1} &\quad {\mbox{if } \phi \in \mathcal{O}}\\
            {0} &\quad {\mbox{otherwise.}}
        \end{array}
    \right.
\end{equation}

\noindent The set $\mathcal{O}$ holds the values of $\phi$ for which the models are nested. More formally, for any choice of $\phi \in \mathcal{O}$, model $m$ reduces to $m^\dagger$. Note that the nesting constraint of Eq.~\ref{eq:nesting_constraint} effectively defines the ranges for the integrals in Eq.~\ref{eq:ep_ratio}. 

The encompassing prior approach computes the Bayes Factor for nested models and intuitively measures whether a model with constrained $\phi$ better explains the data. In order to evaluate the BF though Eq.~\ref{eq:ep_ratio}, we first estimate the parameter posterior $p(\phi|D, m^\dagger)$ through message passing. The prior $p(\phi|m^\dagger)$ is chosen by the design engineer. Crucially, it follows that performance evaluation of the hearing loss compensation algorithm can be efficiently evaluated for all nested models (the alternative algorithms) at once through executing a message passing schedule on the factor graph of the generative model.

\subsubsection{Model Comparison Example for Hearing Loss Compensation}
\label{sec:bf_example}

For model comparison, we consider two hearing loss models and infer the BF through message passing. The reference model ($m$) is the generative model as described by Eq.~\ref{eq:full_generative_model}. We define the alternative model ($m^\dagger$) to ignore the gain constraint by choosing $p(g_k|g_{k-1}, \theta^\dagger, m^\dagger) = p(g_k|\theta^\dagger, m^\dagger)$. Because the alternative model imposes no restrictions on the gain evolution, it is expected to be overly simplified. The alternative generative model factorization then becomes

\begin{equation}\label{eq:nested_generative_model}
p(s^n,g^n,\theta^\dagger) = p(\theta^\dagger) \prod_{k=1}^n p(s_k|g_k,\theta^\dagger) \, p(g_k|\theta^\dagger) \,,
\end{equation}

\noindent where we omitted the model selection parameter $m^\dagger$.

The reference model reduces to the alternative model when we constrain $\theta = \{\alpha, \beta, \vartheta, \gamma=0\}$. In this case, the gain constraint in the reference model is rendered ineffective, since the low precision allows any gain transition. In practice, this limit leads to a singularity ($0/0$) when computing Eq.~\ref{eq:ep_ratio}. Therefore we choose an upper limit on the gain evolution precision $\gamma$, below which we consider the gain constraint \emph{practically} ineffective. We designate this upper limit with $\omega$. Formally, we define the nesting parameter $\phi = \gamma$ and the nested set $\mathcal{O} = [0, \omega]$. The factor graph and message passing schedule for estimating the posterior over $\gamma$ were already shown in Fig.\ref{fig:fitting_vmp}.

For pedagogical reasons, this paper considers an overly simplified alternative model choice. However, the encompassing prior approach also allows comparisons with model extensions. A design engineer might for example substitute a more complicated hearing loss function, provided that the two models can be rendered equal through a nesting constraint.

\section{Simulations}
\label{sec:simulations}

In order to demonstrate some properties of the proposed design method, we simulated the various inference tasks. These simulations were performed using a custom-built FFG message passing framework, written in the Julia language \cite{bezanson_julia:_2014}, which is a fast high-level language for scientific programming. We used the generative model as specified by the set of equations \ref{eq:full_generative_model}, \ref{eq:observation_model}, \ref{eq:Zurek_model}, \ref{eq:gain_change_prior} and \ref{eq:par-transition-is-delta}.

\subsection{Signal processing}
\label{sec:SP_results}

\begin{figure}[ht]
    \centering
    \includegraphics[width=8cm]{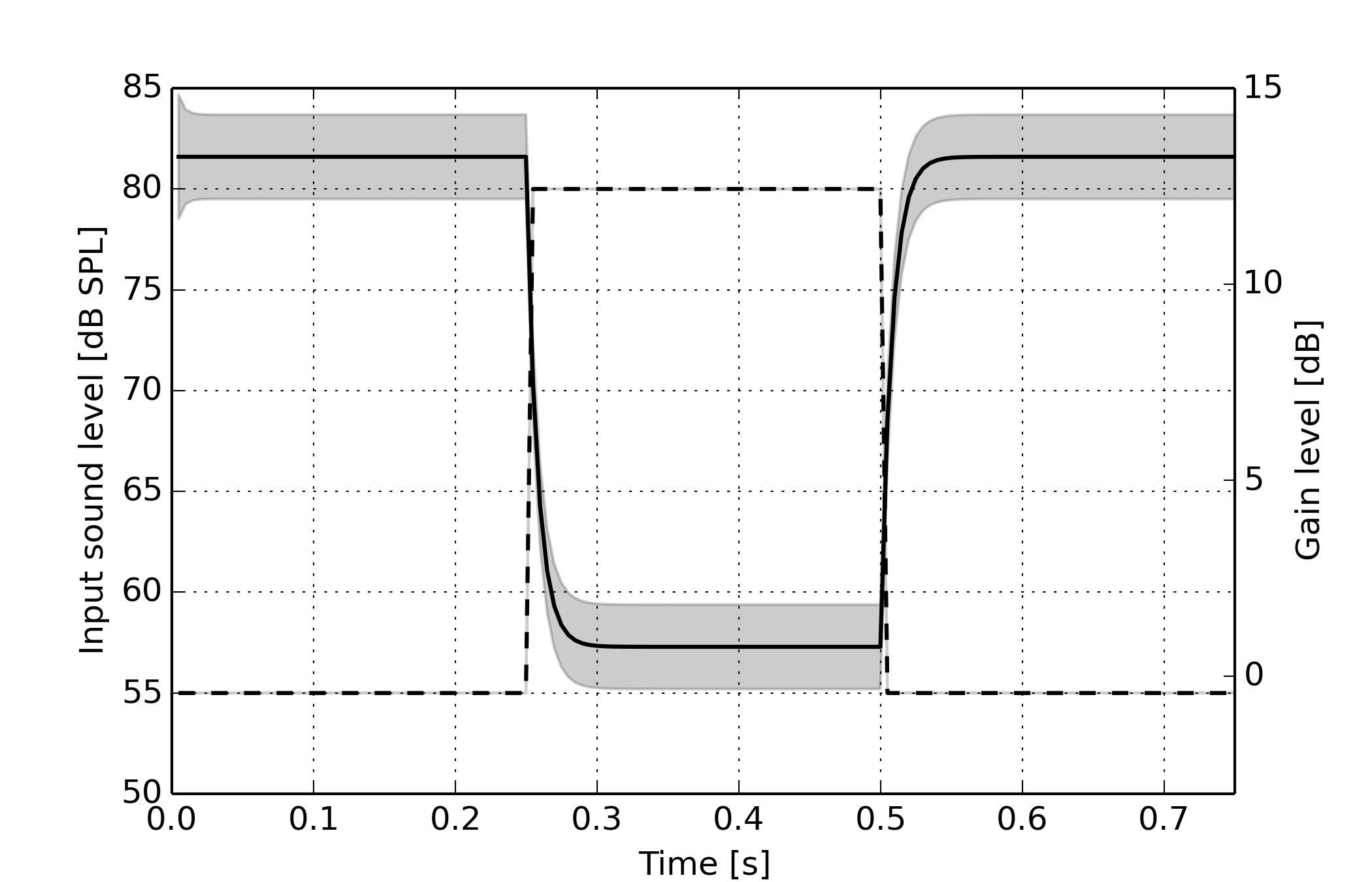}
    \caption{Log-power input (dashed) and inferred log-gain (solid) levels over time for the SP example of Sec.~\ref{sec:SP_results}. The ribbon indicates a confidence interval of one standard deviation.}
    \label{fig:dynamic_gain}
\end{figure}

Zurek's hearing loss model of Eq.~\ref{eq:Zurek_model} was selected with fixed parameter values $\alpha = 2$ and $\beta = -90$, see Fig.~\ref{fig:hl_model}. Furthermore, the variance parameters were fixed to $\vartheta = 10$ and $\gamma = 1$. The log-power signal $s_k$ was assumed to be observed and set to alternating input levels between 80 and 55 [dB SPL], see Fig.~\ref{fig:dynamic_gain}.

\begin{figure}[ht!]
    \centering
    \includegraphics[width=8cm]{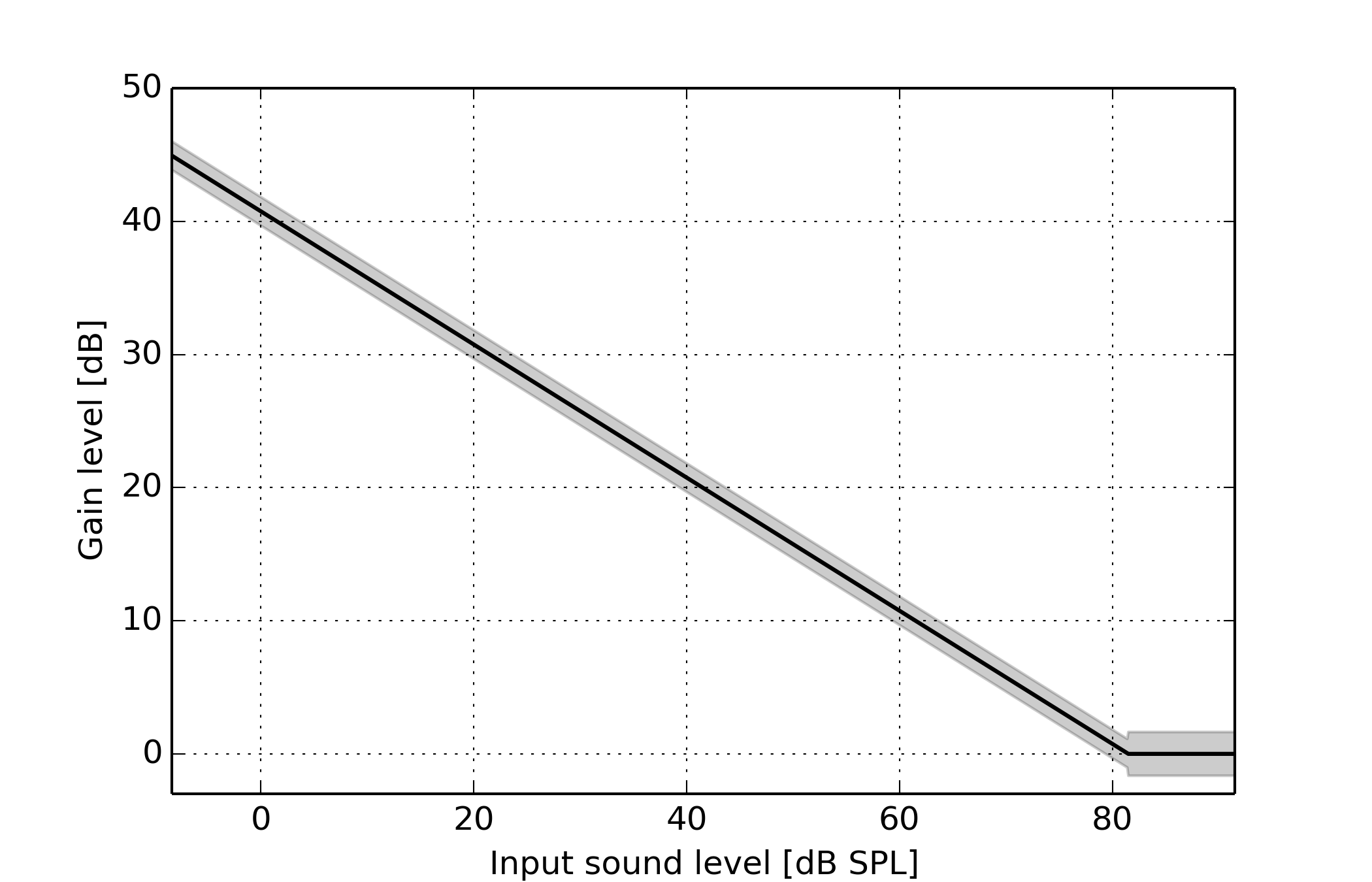}
    \caption{The steady-state input-gain curve for the SP example of Sec.~\ref{sec:SP_results}. The ribbon represents a confidence interval of one standard deviation.}
    \label{fig:static_gain}
\end{figure}

Next, we ran the signal processing message passing scheme of Fig.~\ref{fig:sp-filtering} to infer the gain sequence $g_k$. We show both steady-state and dynamic behavior of $g_k$ in Figs.~\ref{fig:static_gain} and \ref{fig:dynamic_gain}, respectively.

First note that the inferred gain values were modeled by a mean and standard deviation of a Gaussian. Any uncertainty about both gain transitions ($\gamma$) and observations ($\vartheta$) as declared by the generative model is ultimately transferred to uncertainty about the inferred gain. In a full Bayesian treatment, the impact of the gain uncertainty can be properly marginalized when computing the final audio output by 

$$
p(y_k|s^k) = \int p(y_k|g_k)\,p(g_k|s^k)\,\d{g_k} \,.
$$ 

In a practical algorithm, we could simply take the mean of the inferred gain sequence as the outcome of the signal processing stage.

Fig.~\ref{fig:static_gain} reveals that initially the steady-state gain decreases with input power. In the Dynamic Range Compression (DRC) literature it is common to quantify the slope by the \emph{compression ratio} $\mathrm{CR} \triangleq \Delta \text{input} / \Delta (\text{input}+\text{gain})$, which evaluates here to $2:1$. 

Fig.~\ref{fig:dynamic_gain} reveals that the output does not follow the input instantly. As shown in Sec.~\ref{sec:SP_discussion}, our Bayesian signal processing algorithm can also be described as a \emph{Kalman filtering} process. In a Kalman filter, time-constants are represented by a so-called \emph{Kalman gain} ($K_k$ in Eq.~\ref{eq:sp_analytic}). Crucially, the Kalman gain is a function of the variance and precision parameters, $\vartheta$ and $\gamma$, and is automatically updated as part of the signal processing task. As a result, transient behavior in a Bayesian dynamic range compressor is an \emph{emerging} property of uncertainty in our model specification. Intuitively, since we specified the processing goal for the current time step $k$ as an \emph{approximate} loudness restoration task (by a Gaussian model), the system will smooth the loudness restoration task over several time steps. In our simulations, dynamic properties were characterized by measuring the time interval from a step onset to the time that the output signal stabilizes within 2 [dB] of its final value, which is the standard method for HA processing, \cite{dillon_hearing_2012}. Trailing the upward and downward step, these intervals are called the \emph{attack} and \emph{release} time constants respectively. In our simulation, both attack and release time constants evaluate to 15 [ms].   

Altogether, we conclude that the inferred signal processing scheme exhibits the behavior of a Dynamic Range Compressor (DRC) algorithm. Crucially, we did not explicitly design a DRC circuit. Instead, we defined a generative model that included specification of a hearing loss model (which is a problem statement, not a solution), and the DRC behavior is a consequence of the (SP) inference scheme. For a different hearing loss model, a different DRC algorithm would have been inferred. 

\subsection{Collecting training data}
\label{sec:training_data}

For validating simulations of the PE and MC stages, we need a set of preferred examples of signal processing input-output pairs. Here, we chose to generate a training data base by executing the signal processing procedure in a ``preferred'' generative model $m^*$ with hearing loss model $\HL^*$, which is shown as the dash-dotted curve in Fig.~\ref{fig_pe_model_est}. Specifically, the loudness restoration and gain evolution constraints were specified (respectively) as
 
\begin{subequations}
\label{eq:oracle}
\begin{align}
    p(s_k | g_k, \HL^*, m^*) &= \delta(s_k - \HL^*(s_k + g_k)) \\
    p(g_k | g_{k-1}, m^*) &= \mathcal{N}(g_k|g_{k-1},\infty)\,.
\end{align} 
\end{subequations}

We executed the SP task in the corresponding generative model and collected the input-output sequences in a training data base $D=\{\tilde{s}, \tilde{g}\}$. 

\begin{figure}[ht]
    \centering
    \includegraphics[width=8cm]{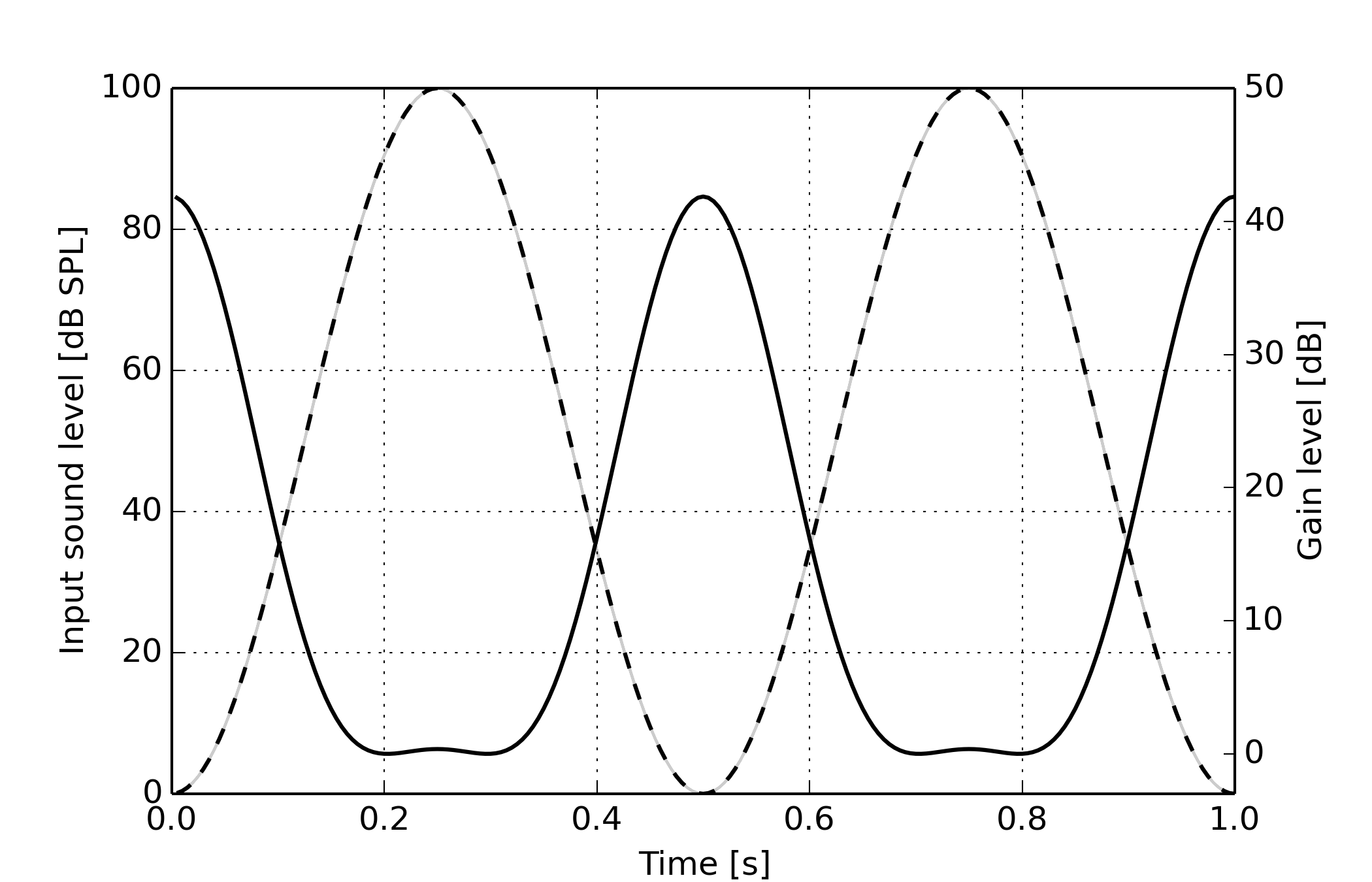}
    \caption{Training data containing selected input $\tilde{s}$ (dashed) and gain $\tilde{g}$ (solid).}
    \label{fig_mc_train}
\end{figure}

\subsection{Parameter estimation}
\label{sec:PE_results}

The PE simulation in this section shows how parameters for Zurek's piecewise loudness growth curve can be estimated from the training data. As a ``learning'' (recognizing) model, we chose Zurek's piecewise linear functional form (Eq.~\ref{eq:Zurek_model}). Prior distributions for the slope and offset were chosen to be constrained to plausible values within the context of Zurek's model: $p(\alpha) = \N{\mu_{\alpha}=1.5, \vartheta_{\alpha}=0.2}$; $p(\beta) = \N{\mu_{\beta}=-50, \vartheta_{\beta}=100}$ respectively. 

The prior distributions for variance and precision for the observation model and the precision for the gain constraint were chosen as $p(\vartheta) = \Ig{a_\vartheta=12, b_\vartheta=110}$ and $p(\gamma) = \Gam{a_\gamma=10, b_\gamma=1}$ respectively.

\begin{figure}[ht]
    \centering
    \includegraphics[width=8cm]{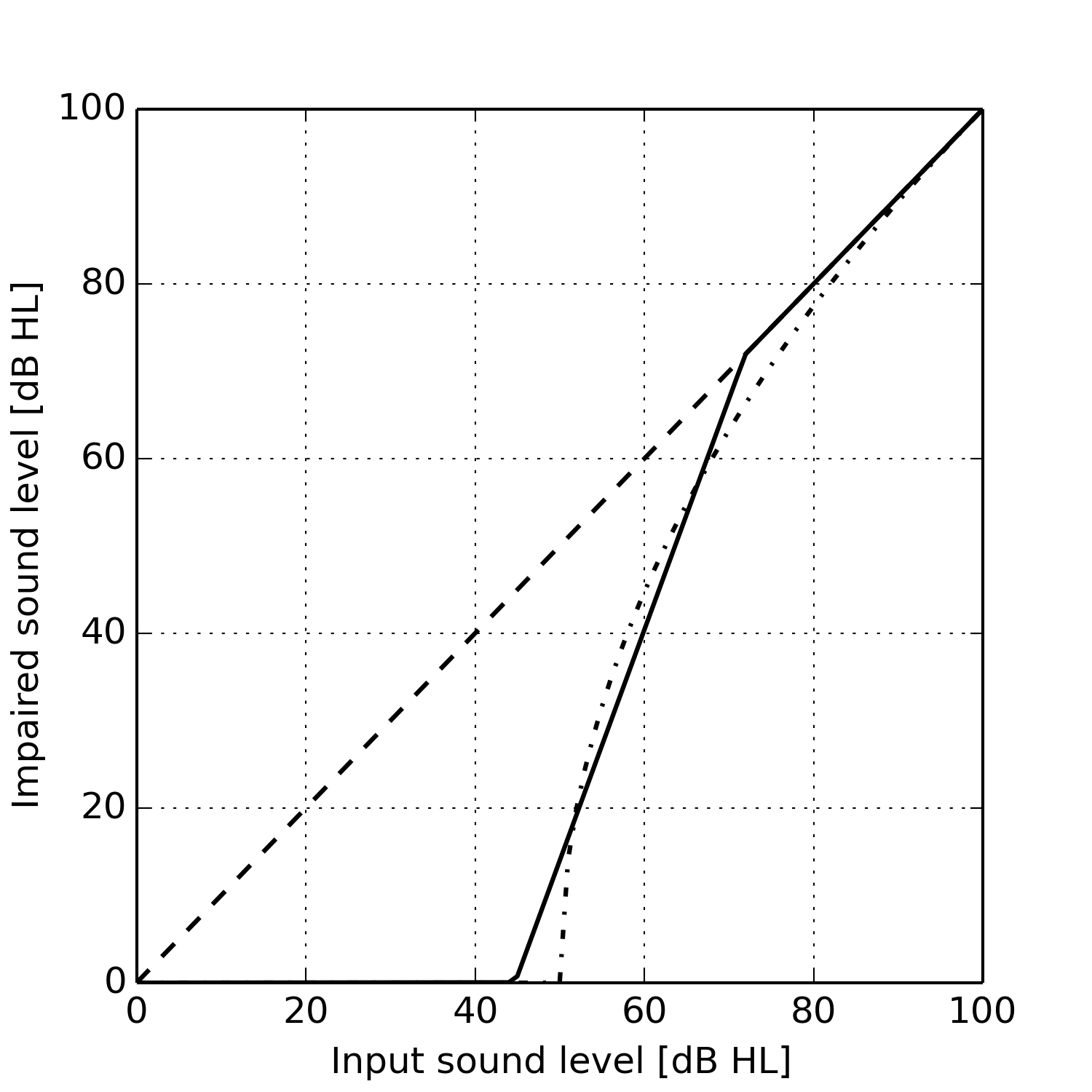}
    \caption{Parameter estimation results for the simulation in Sec.~\ref{sec:PE_results}. Unimpaired (dashed), target (dash-dotted), and estimated (solid) loudness recruitment curves. Hearing loss parameters were estimated at (estimate $\pm$ sd): $\alpha = 2.642 \pm 0.001$ and $\beta = -118.14 \pm 0.03$. The noise variances were estimated at $\vartheta = 9.87$ with variance $0.89$ and $\gamma = 0.94$ with variance $0.008$.}
    \label{fig_pe_model_est}
\end{figure}

Parameter estimation was performed by executing the schedules in Fig.~\ref{fig:fitting_vmp} for 200 iterations. A visual inspection reveals a pleasing similarity between the estimated and target hearing loss models, see Fig.~\ref{fig_pe_model_est}.

In a real-world application, the target model (including the dash-dotted curve in Fig.~\ref{fig_pe_model_est}) is not available. Therefore, it would make sense to search for alternative hearing loss models that provide a better fit to the training data. In the next section we use the Bayes factor to compare the present model against an alternative.

\subsection{Model comparison}
\label{sec:MC_results}

We evaluated the Bayes factor (BF) in order to compare the performance of the reference model (Eq.~\ref{eq:full_generative_model}) against an alternative model (Eq.~\ref{eq:nested_generative_model}) where we omitted the gain contraint. We performed the MC simulation with respect to the training data $\{\tilde{s}, \tilde{g}\}$ generated in Sec.~\ref{sec:training_data}. The BF was calculated according to the procedure outlined in Sec.~\ref{sec:bf_example}. The upper limit for the nesting constraint was chosen at $\omega=0.25$.

The resulting Bayes factor, $\mathrm{BF} = -16.7$ [dHart], rules strongly in favor of the reference model (the model \emph{with} the gain constraint).

\section{Related Work}

The state-of-the-art in hearing aid signal processing is well described by Kates \cite{kates_digital_2008} and Hamacher \cite{hamacher_signal_2005}. More specifically, the literature on dynamic range compression technology for hearing loss compensation is nicely summarized by \cite{dillon_hearing_2012} and \cite{kates_understanding_2010}. In both works, DRC circuits are developed through direct design, i.e., the hearing loss problem is not an explicit part of the solution. In contrast, a problem-based signal processing solution for hearing loss compensation has first been formulated in \cite{farmani_probabilistic_2014}, where an optimal compensation gain is computed through Kalman filtering. The current paper extends that work by proposing a fully probabilistic modelling approach for both the SP, PE and MC tasks as well as an in-situ executable data base collection method. Moreover, the current work presents a factor graph framework for efficient execution of these tasks through message passing on FFGs. 

We used both sum-product and variational message passing to execute the SP, PE and MC tasks. We focused on the design framework and due to space limitations we did not include a detailed derivation of each message in the graphs. Our message update rules are fairly standard. A detailed account of inference through sum-product message passing on FFGs is discussed in \cite{kschischang_factor_2001}, \cite{loeliger_introduction_2004, loeliger_factor_2007} and in the dissertation by Korl \cite{korl_factor_2005}. Variational message passing in FFGs has been described in \cite{dauwels_variational_2007} and \cite{beal_variational_2003}.

The fully probabilistic treatment of hearing loss compensation that is proposed in this work is to our knowledge new to the hearing aid literature. However, the idea of inferring audio processing algorithms though inference in a generative probabilistic model goes back at least to Roweis \cite{roweis_unifying_1999}. More recently, Rennie and colleagues have described several audio processing algorithms for speech recognition and source separation based on probabilistic inference through message passing in a graphical model \cite{rennie_single-channel_2009}. More generally, algorithm design based on inference in generative probabilistic models is an increasingly popular technique in the Bayesian machine learning literature, e.g. \cite{bishop_pattern_2006}, \cite{murphy_machine_2012}.  

The encompassing prior approach to nested model comparison is described in \cite{klugkist_bayesian_2005}. More discussion on Bayesian methods for comparing constrained models are available in \cite{gelfand_bayesian_1992}, \cite{klugkist_bayes_2007} and \cite{wetzels_encompassing_2010}. 

\section{Discussion}

\subsection{Signal processing as Kalman filtering}
\label{sec:SP_discussion}

In Fig.~\ref{fig:sp-filtering}, a sequence of sum-product messages infers the signal processing algorithm for time step $k$. The actual implementation of the signal processing step on an ultra-low-power HA DSP processor may be computationally optimized. If we write out the message sequence explicitly for Zurek's hearing loss model and eliminate intermediate results, a recursive algorithm for computing $g_k$ at time step $k$ emerges, see Eq.~\ref{eq:sp_analytic}.

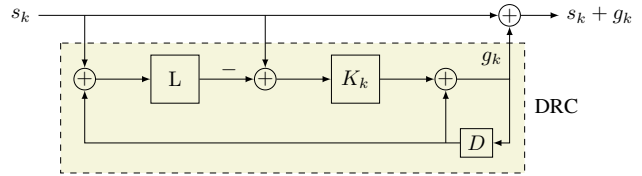
\begin{figure}[ht]
    \centering
    \scalebox{0.8}{
\begin{tikzpicture}
    [node distance=15mm,auto,>=stealth']
    \begin{scope}

        \node[] (s_n) {$s_k$};
        \node[below right of=s_n, roundbox] (l_plus) {$+$};
        \node[box, right of=l_plus] (L) {$\mathrm{L}$};
        \node[right of=L, roundbox] (m_plus) {$+$};
        \node[left of=m_plus, node distance=6mm, yshift=1.5mm] (minus) {$-$};
        \node[box, right of=m_plus] (K) {$K_k$};
        \node[roundbox, right of=K] (r_plus) {$+$};
        \node[roundbox, above right of=r_plus] (t_plus) {$+$};
        \node[box, below right of=r_plus, xshift=-0.55cm, minimum size=3mm] (D) {$D$};
        \node[right of=t_plus] (out) {$s_k + g_k$};

        \path[line] (D) -- ($(l_plus)+(0,-1.061)$) -> (l_plus);
        \path[line] (s_n) -> (t_plus);
        \path[line] ($(l_plus)+(0,1.061)$) -> (l_plus);
        \path[line] ($(m_plus)+(0,1.061)$) -> (m_plus);
        \path[line] ($(r_plus)+(0,-1.061)$) -> (r_plus);
        \path[line] (r_plus) -- ($(r_plus)+(1.061,0)$) -> node[anchor=east, pos=0.4]{$g_k$} (t_plus);
        \path[line] ($(t_plus)+(0,-1.061)$) -- ($(t_plus)+(0,-2.122)$) -> (D);
        \path[line] (l_plus) -> (L);
        \path[line] (L) -> (m_plus);
        \path[line] (m_plus) -> (K);
        \path[line] (K) -> (r_plus);
        \path[line] (t_plus) -> (out);

        \begin{pgfonlayer}{bg}
            \draw[dashed, fill=beige] ($(l_plus)+(-0.4,0.6)$) rectangle ($(D)+(0.8,-0.5)$);
            \node at (8.9,-1.5) {DRC};
        \end{pgfonlayer}
    \end{scope}    
\end{tikzpicture}
    }
    \caption{Signal flow diagram for dynamic range compression by Kalman filtering. The algorithm represented by this flow diagram emerges from the (SP) message passing schedule in Fig.~\ref{fig:sp-filtering}.}
    \label{fig_drc_circuit}
\end{figure}

The algorithm in Eq.~\ref{eq:sp_analytic} describes a dynamic range compressor that follows the structure of a Kalman filter, see Fig.~\ref{fig_drc_circuit}. The signal flow diagram of Fig.~\ref{fig_drc_circuit} and the message passing sequence in Fig.~\ref{fig:sp-filtering} lead to exactly the same posterior for $g_k$. Note that, in contrast to conventional DRCs, the hearing loss problem definition $\HL$ is explicitly included in Fig.~\ref{fig_drc_circuit}. 

The computational cost of the Kalman filter-based SP algorithm is comparable to a regular DRC circuit. In particular if the Kalman filter-based SP algorithms runs independently in each frequency band, the inversion for $K_k$ in Eq.~\ref{eq:sp_analytic} is simply a division.

\begin{align}
\label{eq:sp_analytic}
\text{Assume prior:}& \; p(g_{k-1}|s^{k-1}) = \N{\bar{g}_{k-1}, \vartheta_{g,k-1}} \notag \\
\text{Assume new observation:}& \; s_k \notag \\
\text{Compute:}& \notag \\ 
    a_k &:= \left\{ 
        \begin{array}{c l}
            {\alpha} & \quad {\mbox{if } s_k < \beta/(1-\alpha)} \notag  \\ 
            {1} & \quad {\mbox{otherwise}} 
        \end{array}
    \right.\notag \\
    \vartheta_{u, k} &:= \gamma^{-1}+\vartheta_{g,k-1} \notag \\
    K_k &:= a_k \vartheta_{u, k} \cdot (\vartheta + a_k^2 \vartheta_{u, k})^{-1} \notag \\
    \bar{g}_k &:= \bar{g}_{k-1} + K_k \cdot \left(s_k - \HL\left(s_k+\bar{g}_{k-1}\right) \right)\notag \\
    \vartheta_{g,k} &:= (1 - K_k a_k) \cdot \vartheta_{u, k} \notag \\
    \text{Assign update:}& \; p(g_k|s^k) := \N{\bar{g}_k, \vartheta_{g,k}}\,.
\end{align}

\subsection{Signal processing as a problem inversion process}

In a conventional HA signal processing algorithm, the problem statement (namely: a model for hearing loss) is not an explicit part of the solution (the dynamic range compressing algorithm). Therefore, it is hard to establish which (exact) problem is actually compensated by the signal processing algorithm. In contrast, the proposed method starts with an explicit problem specification (the hearing loss model) and some (possibly vague) constraints on the solution, e.g., a loudness restoration constraint. The set of solution constraints are combined with the problem statement into a single generative probabilistic model. All tasks of interest: signal processing, fitting and performance evaluation, can now be automatically inferred using standard probability theory.      

In our opinion, the inclusion of a problem statement within the signal processing solution contributes to a more principled fitting strategy. A fitting method without an explicitly embedded problem statement is fully dependent on external domain knowledge that is contained in prescriptive fitting rules. In contrast, the Bayesian framework explicitly specifies domain knowledge in the generative model and applies a neutral fitting strategy (Eq.~\ref{eq:pe}) that relies on this internally stored domain knowledge. For instance, complicated knowledge about a dynamic hearing loss feature can be described in the hearing loss model. Therefore, there is no need to translate the available domain knowledge to fitting rules. Furthermore, the Bayesian approach fits \emph{all} HA tuning parameters, including attack and release times. Moreover, the proposed method is easily combined with existing fitting strategies. Any pre-knowledge about optimal tuning parameter settings can simply be incorporated as parameter priors.

The explicit hearing loss problem statement also contributes to our understanding of the performance evaluation task. Conventional (non-clinical) hearing aid algorithm evaluation relies on speech intelligibility and quality metrics. These metrics again incorporate domain knowledge about the speech signal and human auditory system, e.g., AI \cite{pavlovic_articulation_1986}, SII \cite{houtgast_evaluation_1971}, PESQ \cite{beerends_speech_2008}, PEMO-Q \cite{huber_pemo-q_2006} and HASQI \cite{kates_hearing-aid_2010} metrics. The probabilistic approach in this paper conceptually separates the problem definition (generative model) from the performance metric (Bayes Factor). Note that the Bayes Factor follows from the combination of the problem statement (the generative model) and the proposed solution (the fitted HA algorithm). It suffices then to use a neutral (without domain knowledge) performance evaluation metric. In other words, the Bayes Factor simply measures how well the proposed solution solves the stated problem, based on the observed data.  

Finally, we wish to draw attention to the personalization aspects of the proposed design method. In a conventional design approach, the most important measurement on the patient is the hearing threshold (the `audiogram`). It is well-known that the audiogram is an incomplete description of hearing impairment. For instance, the audiogram contains no information about dynamic aspects. The proposed approach facilitates the description of a complicated hearing impairment problem by a dynamic model. All parameters are then tuned by personally selected HA processing examples. This personalization process could be a fine-tuning process after a professional audiologist (or prescriptive rules) has set appropriate priors on the tuning parameter distributions, so as to warrant the desired audiological effect of the hearing aid. 

\subsection{Further research}

We regard the proposed hearing aid design method in this paper as a first step toward in-situ personalizable hearing aid algorithms. Clearly, more work is required before the proposed method is suited for industrial applications. This paper has presented a new design framework and tried to point to potential advantages of the proposed method. Furthermore, the method was analyzed through simulations. We emphasize that the presented hearing loss compensation algorithm served as an example to clarify the design process, rather than an effort to outperform existing dynamic range compressors. 

From a practical perspective, an important question remains. Namely, which hearing loss models actually lead (after Bayesian inversion) to pleasing SP algorithms. In a sense, any SP algorithm corresponds to an explicitly stated hearing loss model. The proposed method then accomplishes a very important aspect of hearing science, namely the discovery of a proper hearing loss model. Clearly, an extensive clinical evaluation would be required before these ideas could be adopted in a clinical practice. 

\section{Conclusions}

In this paper we have addressed a personalized hearing aid design method. We described in detail the signal processing (SP), parameter estimation (PE) and model comparison (MC) tasks for hearing aid algorithm design by probabilistic inference on a generative model for hearing loss compensation. All inference methods are executed as message passing schemes on Forney-style factor graphs, thus facilitating automated derivation of the SP, PE and MC tasks. The factor graph framework potentially allows for realization on a hearing aid or mobile device hardware. For Zurek's hearing loss model, we showed that the inferred SP algorithm corresponds to a dynamic range compressor. Furthermore, all parameters in the generative model can be tuned by an appropriate variational message passing scheme in the FFG of the generative model. In this framework, there is no need for prescriptive fitting rules. The proposed Bayesian model comparison procedure provides a principled method for choosing a best algorithm structure amongst alternatives, based on a user-selected data set. For a nested family of model candidates, MC can efficiently be executed through the encompassing prior approach.

\appendices
\section*{Acknowledgment}
The authors gratefully acknowledge stimulating discussions with Marco Cox, Tjalling Tjalkens and Ren\'{e} Duijkers of the signal processing systems group at TU/e and with Joris Kraak of GN ReSound. The authors also thank three anonymous reviewers for their valuable comments.

\bibliographystyle{IEEEtran}
\bibliography{bibliography}

\vspace{-1.0cm}

\begin{IEEEbiography}[{\includegraphics[width=1in,height=1.25in,clip,keepaspectratio]{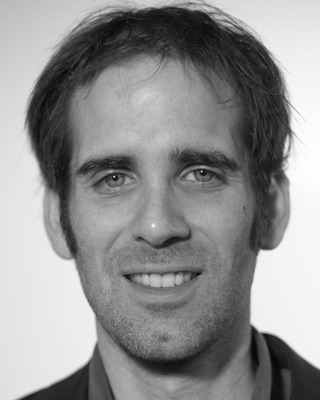}}]{Thijs van de Laar}
received the M.Sc. degree in natural sciences from the Radboud University Nijmegen in 2010. Currently he works as a Ph.D. student in the signal processing systems group at Eindhoven University of Technology. His research focusses on personalization of audio processing algorithms through the development and application of modern machine learning techniques.
\end{IEEEbiography}

\vspace{-1.0cm}

\begin{IEEEbiography}[{\includegraphics[width=1in,height=1.25in,clip,keepaspectratio]{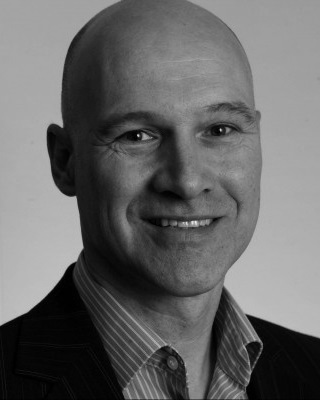}}]{Bert de Vries}
received M.Sc. (1986) and Ph.D. (1991) degrees in Electrical Engineering from Eindhoven University of Technology (TU/e) and the University of Florida, respectively. From 1992 until 1999 he worked at Sarnoff Research Center in Princeton (NJ), where he contributed to research projects over a wide range of signal and image processing topics. Since April 1999 he has been employed in the hearing aids industry (currently at GN ReSound), both as a Principal Scientist and as a Research Manager. Since January 2012 he is also a Professor at the Signal Processing Systems Group at TU/e, where he conducts research on developing Bayesian machine learning techniques for efficient tuning of hearing aids to the personal preferences of end users. 
\end{IEEEbiography}

\end{document}